\pdfoutput=1
\documentclass[11pt]{article}

\usepackage[final]{naacl2021}
\usepackage{times}
\usepackage{latexsym}

\usepackage{nert}

\usepackage[T1]{fontenc}
\usepackage[utf8x]{inputenc}

\usepackage{microtype}

\usepackage{listings}
\usepackage{hyperref}
\usepackage{tablefootnote}

\usepackage{pifont}% http://ctan.org/pkg/pifont

\usepackage{amsmath}
\usepackage{verbatim}
\usepackage{textcomp} 
\usepackage{enumitem} 

\usepackage{multirow}
\usepackage{caption}
\usepackage{subcaption}

\newcommand{\bleu}{{\textsc{Bleu}}}

\newcommand{\chrf}[1]{\textsc{ChrF}{$_{#1}$}}

\newcommand{\maf}[1]{\textsc{MacroF}{\text{$_{{#1}}$}}}
\newcommand{\mif}[1]{\textsc{MicroF}{\text{$_{{#1}}$}}}

\newcommand{\blrtmn}{\textsc{BLEURT}\text{\footnotesize{}mean}}
\newcommand{\blrtmd}{\textsc{BLEURT}\text{\footnotesize{}median}}

\newcommand{\insig}{$^{\times}$}

\usepackage{amssymb}

\usepackage{tikz}
\usepackage[normalem]{ulem}
\usetikzlibrary{positioning,fit,calc,backgrounds}
\tikzset{block/.style={draw,thick,text width=2cm,minimum height=1cm,align=center},
         line/.style={-latex}
}
\makeatletter
\def\ruwave{\bgroup \markoverwith{\lower3.5\p@\hbox{\textcolor{red}{\sixly \char58}}}\ULon}
\font\sixly=lasy6
\makeatother

\usepackage{graphicx,wasysym}

\title{Macro-Average: Rare Types Are Important Too}

 \author{
    Thamme Gowda \\ 
    Information Sciences Institute \\
    University of Southern California\\
    \eml{tg@isi.edu}
 \And 
    Weiqiu You \\
    Dept of Computer and Information Science \\
    University of Pennsylvania \\
    \eml{weiqiuy@seas.upenn.edu}
 \AND
    Constantine Lignos \\
    Michtom School of Computer Science \\
    Brandeis University \\
    \eml{lignos@brandeis.edu}
  \And 
    Jonathan May \\
    Information Sciences Institute \\
    University of Southern California \\
    \eml{jonmay@isi.edu}
        }

\date{}

\begin{document}
\maketitle
\begin{abstract}
  While traditional corpus-level evaluation metrics for machine translation~(MT) correlate well with fluency, they struggle to reflect adequacy.
Model-based MT metrics trained on segment-level human judgments have emerged as an attractive replacement due to strong correlation results.
These models, however, require potentially expensive re-training for new domains and languages.
Furthermore, their decisions are inherently non-transparent and appear to reflect unwelcome biases. 
We explore the simple type-based classifier metric, \maf1, and study its applicability to MT evaluation. 
We find that \maf1 is competitive on direct assessment, and outperforms others in indicating downstream cross-lingual information retrieval task performance.
Further, we show that \maf1 can be used to effectively compare supervised and unsupervised neural machine translation, and reveal significant qualitative differences in the methods' outputs.\footnote{Tools and analysis are available at \url{https://github.com/thammegowda/007-mt-eval-macro}.
MT evaluation metrics are at \url{https://github.com/isi-nlp/sacrebleu/tree/macroavg-naacl21}.
}

%\myurl{https://github.com/thammegowda/007-mt-eval-macro}

\end{abstract}

\section{Introduction}

Model-based metrics for evaluating machine translation such as BLEURT \cite{sellam-etal-2020-bleurt}, ESIM \cite{mathur-etal-2019-ESIM}, and YiSi \cite{lo-2019-yisi} have recently attracted attention due to their superior correlation with human judgments \cite{WMT19-metrics-proceedings}. However, \bleu~\cite{papineni-etal-2002-bleu} remains  the most widely used corpus-level MT metric. It correlates reasonably well with human judgments, and moreover is easy to understand and cheap to calculate, requiring only reference translations in the target language. By contrast, model-based metrics require tuning on thousands of examples of human evaluation for every new target language or domain \cite{sellam-etal-2020-bleurt}. Model-based metric scores are also opaque and can hide undesirable biases, as can be seen in Table~\ref{tab:bleurt-bias}.

\begin{table}[ht]
    \centering
    \footnotesize
    \begin{tabular}{l l l }
Reference:& \multicolumn{2}{l}{You must be a doctor.} \\
Hypothesis: & \multicolumn{2}{l}{$\rule{1cm}{0.15mm}$ must be a doctor.} \\
    % & You &	~0.990 \\
    & He	&-0.735 \\
    %& Alexandra & -0.888 \\
    %& Alexander & -0.975 \\
    & Joe & -0.975 \\
    & Sue & -1.043 \\
    & She	 &-1.100 \\\hline
Reference:& \multicolumn{2}{l}{It is the greatest country in the world.} \\
Hypothesis:& \multicolumn{2}{l}{$\rule{1cm}{0.15mm}$ is the greatest country in the world.} \\
    % & It	& ~0.957 \\
    & France &	-0.022 \\
    & America	& -0.060 \\
    & Russia &	-0.161 \\
    %& China  & -0.166 \\
    %& USA    & -0.168 \\
    %& India   &	-0.211 \\
    & Canada  & -0.309 
    \end{tabular}
    \caption{A demonstration of BLEURT's internal biases; model-free metrics like BLEU would consider each of the errors above to be equally wrong.}
    \label{tab:bleurt-bias}
\end{table}

The source of model-based metrics' (e.g. BLEURT) correlative superiority over model-free metrics (e.g. BLEU) appears to be the former's ability to focus evaluation on \textit{adequacy}, while the latter are overly focused on \textit{fluency}. BLEU and most other generation metrics consider each output \textit{token} equally. Since natural language is dominated by a few high-count types, an MT model that concentrates on getting its \textit{if}s, \textit{and}s and \textit{but}s right will benefit from BLEU in the long run more than one that gets its \textit{xylophone}s, \textit{peripatetic}s, and \textit{defenestrate}s right. Can we derive a metric with the discriminating power of BLEURT that does not share its bias or expense and is as interpretable as BLEU? 

As it turns out, the metric may already exist and be in common use. Information extraction and other areas concerned with classification have long used both \textit{micro averaging}, which treats each token equally, and \textit{macro averaging}, which instead treats each \textit{type} equally, when evaluating. The latter in particular is useful when seeking to avoid results dominated by overly frequent types.  In this work we take a classification-based approach to evaluating machine translation in order to obtain an easy-to-calculate metric that focuses on adequacy as much as BLEURT but does not have the expensive overhead, opacity, or bias of model-based methods.

Our contributions are as follows:
We consider MT as a classification task, and thus admit \maf1 as a legitimate approach to evaluation~(Section ~\ref{sec:mt-as-cls}). 
We show that \maf1 is competitive with other popular methods at tracking human judgments in translation (Section~\ref{sec:wmt-metrics}). 
We offer an additional justification of \maf1 as a performance indicator on adequacy-focused downstream tasks such as cross-lingual information retrieval (Section \ref{sec:clir}). 
Finally, we demonstrate that \maf1 is just as good as the expensive BLEURT at discriminating between structurally different MT approaches in a way \bleu\ cannot, especially regarding the adequacy of generated text, and provide a novel approach to qualitative analysis of the effect of metrics choice on quantitative evaluation (Section \ref{sec:unmt}).

\section{NMT as Classification}
\label{sec:mt-as-cls}
Neural machine translation (NMT) models are often viewed as pairs of encoder-decoder networks.
Viewing NMT as such is useful in practice for implementation; however, such a view is inadequate for theoretical analysis. 
\citet{gowda2020finding} provide a high-level view of NMT as two fundamental ML components: an autoregressor and a classifier. 
Specifically, NMT is viewed as a multi-class classifier that operates on representations from an autoregressor.
We may thus consider classifier-based evaluation metrics.

Consider a test corpus, $T = \{ (x^{(i)}, h^{(i)}, y^{(i)}) | i = 1,2,3...m \}$ where $x^{(i)}$, $h^{(i)}$, and $y^{(i)}$ are source, system hypothesis, and reference translation, respectively. Let $x = \{x^{(i)} \forall i\}$ and similar for $h$ and $y$.  Let $V_h, V_y, V_{h\cap y},$ and $V$ be the vocabulary of $h$, the vocabulary of $y$, $V_h \cap V_y$, and $V_h \cup V_y$, respectively.
For each class $c \in V$, 
\begin{align*}
 \textsc{Preds}(c) &= \sum_{i=1}^m C(c, h^{(i)})\\
 \textsc{Refs}(c) &= \sum_{i=1}^m C(c, y^{(i)})\\
\textsc{Match}(c) &= \sum_{i=1}^m min\{C(c, h^{(i)}), C(c, y^{(i)})\} 
\end{align*}
\noindent where $C(c, a)$  counts the number of tokens of type $c$ in sequence $a$~\cite{papineni-etal-2002-bleu}. 
For each class $c \in V_{h \cap y}$, precision ($P_c$), recall ($R_c$), and $F_\beta$ measure ($F_{\beta;c}$) are computed as follows:\footnote{We consider $F_{\beta;c}$ for $c \not\in V_{h \cap y}$ to be 0.}
\begin{align*}
    P_c &= \frac{\textsc{Match}(c)}{\textsc{Preds}(c)} ; \hspace{5mm} R_c = \frac{\textsc{Match}(c)}{\textsc{Refs}(c)} \\
    F_{\beta;c} &= (1 + \beta^2)  \frac{ P_c \times R_c}{ \beta^2 \times P_c + R_c}
\end{align*}

The \textit{macro-average} consolidates individual performance by averaging by type, while the \textit{micro-average} averages by token:  
\begin{align*}
\maf\beta &= \frac{\sum_{c \in V} F_{\beta;c}} {|V|}\\
\mif\beta &= \frac{\sum_{c\in V} f(c) \times F_{\beta;c}} {\sum_{c'\in V} f(c')}
\end{align*}
\noindent where $f(c) = \textsc{Refs}(c)+k$ for smoothing factor $k$.\footnote{We use $k=1$. When $k \rightarrow \infty, \space \mif\beta \rightarrow \maf\beta. $} We scale $\maf\beta$ and $\mif\beta$ values to percentile, similar to \bleu, for the sake of easier readability.

\section{Justification for \maf1}
\label{sec:justific}

In the following sections, we verify and justify the utility of \maf1 while also offering a comparison with popular alternatives such as \mif1, \bleu, \chrf{1}, and BLEURT.\footnote{\bleu\ and \chrf1 scores reported in this work are computed with \textsc{SacreBleu}; see the Appendix for details.
BLEURT scores are from the \textit{base} model \citep{sellam-etal-2020-bleurt}. We consider two varieties of averaging to obtain a corpus-level metric from the segment-level BLEURT: mean and median of segment-level scores per corpus.
}
We use Kendall's rank correlation coefficient, $\tau$, to compute the association between metrics and human judgments.
Correlations with p-values smaller than $\alpha=0.05$ are considered to be statistically significant.

\subsection{Data-to-Text: WebNLG}
\label{sec:webnlg}

\begin{table}[ht]
    \footnotesize
    \centering
    \begin{tabular}{lrr}
Name & Fluency \& Grammar & Semantics \\ \hline\hline
\bleu\  & \insig.444 & \insig.500 \\
\chrf1 & \insig.278    & .778   \\
\maf1  & \insig.222    & .722   \\
\mif1  & \insig.333    & .611   \\ \hline
\blrtmn & \insig.444   & .833   \\
\blrtmd & .611  & .667   \\
\end{tabular}
    \caption{\small WebNLG data-to-text task: Kendall's $\tau$ between system-level MT metric scores and human judgments.
    Fluency and grammar are correlated identically by all metrics.
    Values that are \textit{not} significant at $\alpha=0.05$ are indicated by \insig{}.}
    \label{tab:webnlg-kendall}
\end{table}

We use the 2017 WebNLG Challenge dataset \cite{gardent2017webNLG-corpus, shimorina2018webnlg-human-eval}\footnote{\url{https://gitlab.com/webnlg/webnlg-human-evaluation}} to analyze the differences between micro- and macro- averaging. 
WebNLG is a task of generating English text for sets of triples extracted from DBPedia.
Human annotations are available for a sample of 223 records each from nine NLG systems.
The human judgments provided have three linguistic aspects---fluency, grammar, and semantics\footnote{Fluency and grammar, which are elicited with nearly identical directions \cite{gardent2017webNLG-corpus}, are identically correlated.}---which enable us to perform a fine grained analysis of our metrics.
We compute Kendall's $\tau$ between metrics and human judgments, which are reported in Table~\ref{tab:webnlg-kendall}.

As seen in Table~\ref{tab:webnlg-kendall}, the metrics exhibit much variance in agreements with human judgments. %Fluency and grammar display same correlations and hence are com  
For instance, \blrtmd\ is the best indicator of fluency and grammar, however \blrtmn\ is best on semantics. 
BLEURT, being a \textit{model-based} measure that is directly trained on human judgments, scores relatively higher than others.
Considering the model-free metrics, \chrf1 does well on semantics but poorly on fluency and grammar compared to \bleu.
Not surprisingly, both \mif1 and \maf1, which rely solely on unigrams, are poor indicators of fluency and grammar compared to \bleu, however \maf1 is clearly a better indicator of semantics than \bleu. 
The discrepancy between \mif1 and \maf1 regarding their agreement with fluency, grammar, and semantics is expected: micro-averaging pays more attention to function words (as they are frequent types) that contribute to fluency and grammar whereas macro-averaging pays relatively more attention to the content words that contribute to semantic adequacy. 

The take away from this analysis is as follows: \maf1 is a strong indicator of semantic adequacy, however, it is a poor indicator of fluency. We recommend using either \maf1 or \chrf1 when semantic adequacy and not fluency is a desired goal.

\subsection{Machine Translation: WMT Metrics}
\label{sec:wmt-metrics}

\begin{table*}[ht!]
    %\footnotesize
    \small
    \centering
    
\begin{tabular}{r r l r r r r r }
Year & Pairs  & & $\star$\bleu\ & \bleu\ & \maf1 & \mif1 & \chrf1 \\ \hline\hline
\multirow{3}{*}{ 2019 } 
    & \multirow{3}{*}{18}
     & Mean   & .751 & .771 & .821 & .818 & .841  \\ 
   & & Median & .782 & .752 & .844 & .844 & .875  \\
   & & Wins   &     3 &     3 &  \textbf{6}    &     3 &   5 \\ \hline
\multirow{3}{*}{ 2018 } 
  & \multirow{3}{*}{14}
   & Mean   & .858 & .857 & .875 & .873 & .902  \\ 
  & & Median & .868 & .868 & .901 & .879 & .919  \\
  & & Wins    &  1  &  2 & 3 &  2 &  \textbf{6}\\ \hline  
\multirow{3}{*}{ 2017 }
   & \multirow{3}{*}{13}
    & Mean   & .752 & .713 & .714 & .742 & .804 \\  
  & & Median & .758 & .733 & .735 & .728 & .791 \\
  & & Wins   & 5 & 4 & 2 & 2 & \textbf{6} \\
\end{tabular}   
\caption{WMT 2017--19 Metrics task: Mean and median Kendall's $\tau$ between MT metrics and human judgments.
Correlations that are not significant at $\alpha=0.05$ are excluded from the calculation of mean, and median, and wins.
See Appendix Tables \ref{tab:wmt19-kendall}, \ref{tab:wmt18-kendall}, and \ref{tab:wmt17-kendall} for full details.
$\star$\bleu\ is pre-computed scores available in the metrics packages.
In 2018 and 2019, both \maf1 and \mif1 outperform \bleu, \maf1 outperforms \mif1.
\chrf1 has strongest mean and median agreements across the years.
Judging based on the number of wins, \maf1 has steady progress over the years, and outperforms others in 2019.
}
\label{tab:wmt-summary}
\end{table*}

In this section, we verify how well the metrics agree with human judgments using Workshop on Machine Translation (WMT) metrics task datasets for 2017--2019~\cite{WMT17-metrics,WMT18-metrics,WMT19-metrics-proceedings}.\footnote{\url{http://www.statmt.org/wmt19/metrics-task.html}}
We first compute scores from each MT metric, and then calculate the correlation $\tau$ with human judgments.

As there are many language pairs and translation directions in each year, we report only the mean and median of $\tau$, and number of wins per metric for each year in Table \ref{tab:wmt-summary}.
We have excluded BLEURT from comparison in this section since the BLEURT models are fine-tuned on the same datasets on which we are evaluating the other methods.\footnote{\url{https://github.com/google-research/bleurt}}
\chrf1 has the strongest mean and median agreement with human judgments across the years.
In 2018 and 2019, both \maf1 and \mif1 mean and median agreements outperform \bleu\, whereas in 2017 \bleu\ was better than \maf1 and \mif1.

As seen in Section~\ref{sec:webnlg}, \maf1 weighs towards semantics whereas \mif1 and \bleu\ weigh towards fluency and grammar.
This indicates that recent MT systems are mostly fluent, and adequacy is the key discriminating factor amongst them.
\bleu\ served well in the early era of statistical MT when fluency was a harder objective. 
Recent advancements in neural MT models such as Transformers \cite{vaswani2017attention} produce fluent outputs, and have brought us to an era where semantic adequacy is the focus.

% save it for the slides
%We envision that the methods such as \maf1 that emphasize the long tail be more successful in the future years, and this vision is inline with \citet{steedman-2008-last}:
%One day, either because of the demise of Moore’slaw, or simply because we have done all the easy stuff, the Long Tail will come back to haunt us.
%\textit{``One day, ... simply because we have done all the easy stuff, the Long Tail will come back to haunt us.''}

\subsection{Cross-Lingual Information Retrieval}
\label{sec:clir}
In this section, we determine correlation between MT metrics and  downstream cross-lingual information retrieval (CLIR) tasks.
CLIR is a kind of information retrieval (IR) task in which documents in one language are retrieved given queries in another~\cite{grefenstette2012CLIR}. 
A practical solution to CLIR is to translate source documents into the query language using an MT model, then use a monolingual IR system to match queries with translated documents. 
Correlation between MT and IR metrics is accomplished in the following steps: 
\begin{enumerate}[noitemsep,topsep=0pt]
 \item Build a set of MT models and measure their performance using MT metrics.
 \item Using each MT model in the set, translate all source documents to the target language, build an IR model, and measure IR performance on translated documents.
 \item For each MT metric, find the correlation between the set of MT scores and their corresponding set of IR scores.
 The MT metric that has a stronger correlation with the IR metric(s) is more useful than the ones with weaker correlations.
\item Repeat the above steps on many languages to verify the generalizability of findings.
\end{enumerate}

An essential resource of this analysis is a dataset with human annotations for computing MT and IR performances.
We conduct experiments on two datasets: firstly, on data from the 2020 workshop on \textit{Cross-Language Search and Summarization of Text and Speech} (CLSSTS) \cite{zavorin-etal-2020-corpora}, and secondly, on data originally from Europarl, prepared by \citet{lignos-etal-2019-MT-IR} (Europarl).

\subsubsection{CLSSTS Datasets}
\label{sec:material}

\begin{table*}[ht]
    \footnotesize
    \begin{tabular}{l l l r r r r r r }
 & Domain & IR Score & \bleu\ & \maf1 & \mif1 & \chrf1 & \blrtmn & \blrtmd \\\hline\hline

\multirow{4}{*}{LT-EN} 
& \multirow{2}{*}{In} 
  & AQWV & .429 & \insig.363 & \textbf{.508} & \insig.385 & .451  & .420 \\
& & MAP  & .495  & .429      & \textbf{.575} & .451       & .473  & .486 \\
& \multirow{2}{*}{In+Ext}
   & AQWV & \insig.345 & \textbf{.527} & .491  & .491 & .491 & .477 \\
&  & MAP  & \insig.273 & \insig\textbf{.455} & \insig.418 & \insig.418 & \insig.418 & \insig.404 \\\hline
\multirow{4}{*}{PS-EN}
  & \multirow{2}{*}{In} 
    & AQWV  & .559 & \textbf{.653} & .574 & .581 & .584 & .581  \\
  & & MAP   & .493 & \textbf{.632} & .487 & .494 & .558 & .554 \\
  & \multirow{2}{*}{In+Ext}
    & AQWV   & .589 & \textbf{.682} & .593 & .583 & .581 & .571 \\
  & & MAP    & .519 & \textbf{.637} & .523 & .482 & .536 & .526 \\\hline
\multirow{4}{*}{BG-EN}
 & \multirow{2}{*}{In} 
    & AQWV   & \insig.455 & \textbf{.550}   & .527  & \insig.382 & \insig.418  & .418 \\ 
 &  & MAP    &  .491      &  \textbf{.661}  & .564  &  .491       & .527       & .527 \\ 
 & \multirow{2}{*}{In+ext}
    & AQWV   & \insig.257 & \textbf{.500}       & \insig.330 & \insig.404 & \insig.367 & \insig.367 \\
 &  & MAP   & \insig.183 & \insig\textbf{.426} & \insig.257 & \insig.330 & \insig.294 & \insig.294 
\end{tabular} 
\caption{CLSSTS CLIR task: Kendall's $\tau$ between IR and MT metrics under study.
The rows with Domain=In are where MT and IR scores are computed on the same set of documents, whereas Domain=In+Ext are where IR scores are computed on a larger set of documents that is a superset of segments on which MT scores are computed.
\textbf{Bold} values are the best correlations achieved in a row-wise setting; values with \insig~ are \textit{not} significant at $\alpha=0.05$.}
\label{tab:material-kendall}
\end{table*}

CLSSTS datasets contain queries in English~(EN), and documents in many source languages along with their human translations, as well as query-document relevance judgments. 
We use three source languages: Lithuanian~(LT), Pashto~(PS), and Bulgarian~(BG).
The performance of this CLIR task is evaluated using two IR measures: Actual Query Weighted Value (AQWV) and Mean Average Precision (MAP).
AQWV\footnote{\href{https://www.nist.gov/system/files/documents/2017/10/26/aqwv\_derivation.pdf}{https://www.nist.gov/system/files/documents-/2017/10/26/aqwv\_derivation.pdf}} is derived from Actual Term Weighted Value (ATWV) metric \cite{wegmann2013ATWV}. 
We use a single CLIR system \cite{boschee-etal-2019-saral} with the same IR settings for all MT models in the set, and measure Kendall's $\tau$ between MT and IR measures.
The results, in Table~\ref{tab:material-kendall}, show that \maf1 is the strongest indicator of CLIR downstream task performance in five out of six settings.
AQWV and MAP have a similar trend in agreement to the MT metrics.
\chrf1 and BLEURT, which are strong contenders when generated text is directly evaluated by humans, do not indicate CLIR task performance as well as \maf1, as CLIR tasks require faithful meaning equivalence across the language boundary, and human translators can mistake fluent output for proper translations \cite{callison-burch-etal-2007-meta}.

\subsubsection{Europarl Datasets}
\label{sec:lignos-etal}

\begin{table}[ht]
    \footnotesize
    \centering
\begin{tabular}{l@{\hspace{1mm}} r@{\hspace{1mm}} r@{\hspace{1mm}} r@{\hspace{1mm}} r@{\hspace{1.2mm}} r@{\hspace{1.2mm}} r}
 & \bleu\ & \maf1 & \mif1 & \chrf1 & $\overline{\text{BT}}$ & $\widetilde{\text{BT}}$ \\ \hline\hline
\multirow{1}{*}{ CS-EN } 
 & .850 & .867 & .850 & .850 & \textbf{.900} & .867 \\ 
\multirow{1}{*}{ DE-EN } 
  & .900 & .900 & .900 & .912 & \textbf{.917} & .900 \\
\end{tabular}  
\caption{Europarl CLIR task: Kendall's $\tau$ between MT metrics and RBO. $\overline{\text{BT}}$ and $\widetilde{\text{BT}}$ are short for \blrtmn\ and \blrtmd. All correlations are significant at $\alpha=0.05$.}
\label{tab:lignos-mtir-kendall} 
\end{table}

We perform a similar analysis to Section \ref{sec:material} but on another cross-lingual task set up by \citet{lignos-etal-2019-MT-IR} for Czech $\rightarrow$ English (CS-EN) and German $\rightarrow$ English (DE-EN), using publicly available data from the Europarl v7 corpus~\cite{koehn2005europarl}. 
This task differs from the CLSSTS task (Section \ref{sec:material}) in several ways.
Firstly, MT metrics are computed on test sets from the news domain, whereas IR metrics are from the Europarl domain. The domains are thus intentionally mismatched between MT and IR tests.
Secondly, since there are no queries specifically created for the Europarl domain, GOV2 TREC topics 701–850 are used as domain-relevant English queries.
And lastly, since there are no query-document relevance human judgments for the chosen query and document sets, the documents retrieved by BM25~\cite{jones2000probabilistic} on the English set for each query are treated as relevant documents for computing the performance of the CS-EN and DE-EN CLIR setup. 
As a result, IR metrics that rely on boolean query-document relevance judgments as ground truth are less informative, and we use Rank-Based Overlap (RBO; $p=0.98$) \cite{webber2010RBO} as our IR metric.

We perform our analysis on the same experiments as \citet{lignos-etal-2019-MT-IR}.\footnote{\url{https://github.com/ConstantineLignos/mt-clir-emnlp-2019}}
NMT models for CS-EN and DE-EN translation are trained using a convolutional NMT architecture \cite{gehring2017cnn} implemented in the FAIRSeq~\cite{ott2019fairseq} toolkit.
For each of CS-EN and DE-EN, a total of 16 NMT models that are based on different quantities of training data and BPE hyperparameter values are used.
The results in Table~\ref{tab:lignos-mtir-kendall} show that BLEURT has the highest correlation in both cases.
Apart from the trained \blrtmd\ metric, \maf1 scores higher than the others on CS-EN, and is competitive on CS-EN. \maf1 is not the metric with highest IR task correlation in this setting, unlike in Section \ref{sec:material}, however it is competitive with \bleu\ and \chrf1, and thus a safe choice as a downstream task performance indicator.

\section{Spotting Qualitative Differences Between Supervised and Unsupervised NMT with \maf1}
\label{sec:unmt}

\begin{table*}[ht!]
\centering
\footnotesize

\begin{tabular}{l @{\hspace{3mm}} r @{\hspace{1.5mm}} r @{\hspace{1.5mm}}r |  r@{\hspace{1.5mm}}r@{\hspace{1.5mm}}r |
  r@{\hspace{1.5mm}}r@{\hspace{1.5mm}}r | r@{\hspace{1.5mm}} r@{\hspace{1.5mm}} r | r@{\hspace{1.5mm}} r@{\hspace{1.5mm}} r | r @{\hspace{1.5mm}} r@{\hspace{1.5mm}} r}
& \multicolumn{3}{c|}{\bleu} & \multicolumn{3}{c|}{ \maf1 } & \multicolumn{3}{c|}{ \mif1 } & \multicolumn{3}{c|}{ \chrf1 } & \multicolumn{3}{c|}{ \blrtmn } & \multicolumn{3}{c}{ \blrtmd } \\ 
& SN & UN & $\Delta$ & SN & UN & $\Delta$ & SN & UN & $\Delta$ & SN & UN & $\Delta$ & SN & UN & $\Delta$ & SN & UN & $\Delta$ \\ \hline \hline
DE-EN & 32.7 & 33.9 & -1.2 & 38.5 & 33.6 & 4.9 & 58.7 & 57.9 &  0.8 & 59.9 & 58.0 &  1.9 & .211 & -.026 & .24 & .285 & .067 & .22 \\
EN-DE & 24.0 & 24.0 &  0.0 & 24.0 & 23.5 & 0.5 & 47.7 & 48.1 & -0.4 & 53.3 & 52.0 &  1.3 &-.134 & -.204 & .07 &-.112 &-.197 & .09 \\
FR-EN & 31.1 & 31.2 & -0.1 & 41.6 & 33.6 & 8.0 & 60.5 & 58.3 &  2.2 & 59.1 & 57.3 &  1.8 & .182 &  .066 & .17 & .243 & .154 & .09 \\
EN-FR & 25.6 & 27.1 & -1.5 & 31.9 & 27.3 & 4.6 & 53.0 & 52.3 &  0.7 & 56.0 & 57.7 & -1.7 & .104 &  .042 & .06 & .096 & .063 & .03 \\
RO-EN & 30.8 & 29.6 &  1.2 & 40.3 & 33.0 & 7.3 & 59.8 & 56.5 &  3.3 & 58.0 & 54.7 &  3.3 & .004 & -.058 & .06 & .045 & -.004 & .04 \\
EN-RO & 31.2 & 31.0 &  0.2 & 34.6 & 31.0 & 3.6 & 55.4 & 53.4 &  2.0 & 59.3 & 56.7 &  2.6 & .030 & -.046 & .08 & .027 & -.038 & .07 \\
\end{tabular} 

\caption{For each language direction, UNMT (UN) models have similar \bleu\ to SNMT (SN) models, and \chrf1 and \mif1 have small differences. 
However, \maf1 scores differ significantly, consistently in favor of SNMT. 
Both corpus-level interpretations of BLEURT support the trend reflected by \maf1, but the value differences are difficult to interpret.
}
\label{tab:unmt_vs_snmt}
\end{table*}

\begin{table*}[ht]
    \centering
    \footnotesize
    \begin{tabular}{r @{\hspace{2mm}} l @{\hspace{2mm}} p{0.36\linewidth} | r @{\hspace{2mm}} l @{\hspace{2mm}} p{0.36\linewidth} }
 $\delta_{\maf1}$ & Fav & Analysis 
    & $\delta_{\bleu}$ & Fav   & Analysis \\ \hline \hline
 
 .071   & S  & S: synonym; U: \textit{untranslation}, \textit{noun} 
    &  .048   & S  & S: word order; U: word order, \textit{untranslation}, \textit{ending} \\
 
 .064   & S  & S: synonym; U: \textit{untranslation} 
    & .046   & S  & S: spelling variation; U: synonym, word order, punctuation \\ 
 
 -.055  & U  & U: no issues; S: \textit{untranslation}    
    & .044   & S  & S: extra determiner; U: paraphrase, synonym, \textit{number}, \textit{untranslation}  \\

 .052   & S  & S: synonym; U: \textit{untranslation}, \textit{noun} 
    & .042   & S  & S: synonym; U: synonym, punctuation, extra adverb \\ 
 
 -.045  & U  & U: no issues; S: \textit{untranslation}  
  & -.039  & U  & U: no issues; S: \textit{noun}, \textit{verb}  \\
 
 .044   & S  & S: synonym,  word order; U: \textit{subject}, \textit{truncation}, word order
  &  -.037  & U  & U: no issues; S: punctuation  \\
 
 .044   & S  & S: synonym, tense; U: \textit{untranslation} 
  & -.034  & U  & U: no issues; S: symbol  \\
 
 .043  & S  & S: inflection, word order; U: \textit{ number} 
    & -.032  & U  & U: no issues; S: \textit{adjective}, \textit{noun} \\
 
 -.041  & U  &  U: \textit{adjective}, \textit{verb}; S: \textit{omitted verb}, \textit{untranslation}
  & -.032  & U  & U: \textit{untranslation}; S: tense, \textit{word order}, \textit{meaning}, active/passive voice  \\
   
.041    & S & S: \textit{time}, word order; U: \textit{time}, \textit{nouns}  
  & -.031  & U  & U: \textit{untranslation}; S: word order, synonym, \textit{extra\_conj}  
\end{tabular}
\caption{Analysis of the ten DE-EN test set segments with the most favoritism in SNMT (S) or UNMT (U), according to $\maf1$ (left) and $\bleu$ (right). Fav is the favored system by metrics. The complete text of the sentences is in the Appendix, Tables~\ref{tab:maf1-top-10} and \ref{tab:bleu-top-10}.}
\label{tab:snmt_better_mf1}
\end{table*}

Unsupervised neural machine translation (UNMT) systems trained on massive monolingual data without parallel corpora have made significant progress recently \cite{Artetxe-2018-unmt-iclr,Lample-2018-unmt-iclr,lample-etal-2018-phrase-unmt,conneau-NIPS2019-xlm,Song-2019-MASS,liu2020mbart}. 
In some cases, UNMT yields a \bleu\ score that is comparable with strong\footnote{though not, generally, the strongest} supervised neural machine translation (SNMT) systems. In this section we leverage \maf1 to investigate differences in the translations from UNMT and SNMT systems that have similar \bleu.

We compare UNMT and SNMT for English $\leftrightarrow$ German (EN-DE, DE-EN), English $\leftrightarrow$ French (EN-FR, FR-EN), and English $\leftrightarrow$ Romanian (EN-RO, RO-EN).
All our UNMT models are based on XLM \citep{conneau-NIPS2019-xlm}, pretrained by \citet{XLM-UNMT-Models20}. 
We choose SNMT models with similar \bleu\ on common test sets by either selecting from systems submitted to previous WMT News Translation shared tasks~\cite{bojar-EtAl:2014:W14-33,bojar-EtAl:2016:WMT1} or by building such systems.\footnote{We were unable to find EN-DE and DE-EN systems with comparable \bleu\ in WMT submissions so we built standard Transformer-base~\cite{vaswani2017attention} models for these using appropriate quantity of training data to reach the desired \bleu\ performance. We report EN-RO results with diacritic removed to match the output of UNMT.} Specific SNMT models chosen are in the Appendix (Table~\ref{tab:unmt_vs_snmt2}).

 Table~\ref{tab:unmt_vs_snmt} shows performance for these three language pairs using a variety of metrics. Despite comparable scores in \bleu\ and only minor differences in \mif1 and \chrf1, SNMT models have consistently higher \maf1 and BLEURT than the UNMT models for all six translation directions. 
 
 In the following section, we use a pairwise maximum difference discriminator approach to compare corpus-level metrics \bleu\ and \maf1 on a segment level. Qualitatively, we take a closer look at the behavior of the two metrics when comparing a translation with altered meaning to a translation with differing word choices using the metric.

\subsection{Pairwise Maximum Difference Discriminator}

We consider cases where a metric has a strong opinion of one translation system over another, and analyze whether the opinion is well justified. In order to obtain this analysis, we employ a pairwise segment-level discriminator from within a corpus-level metric, which we call \textit{favoritism}.

We extend the definition of $T$ from Section~\ref{sec:mt-as-cls} to $T = \{ x, h_S, h_U, y\}$ where each of $h_S$ and $h_U$ is a separate system's hypothesis set for $x$.\footnote{The subscripts represent SNMT and UNMT in this case, though the definition is general.}
Let $M$ be a corpus-level measure such that $M(h, y) \in \mathbb{R}$ and a higher value implies better translation quality. $M(h^{(-i)}, y^{(-i)})$ is the corpus-level score obtained by excluding $h^{(i)}$ and $y^{(i)}$ from $h$ and $y$, respectively. We define the \textit{benefit} of segment $i$, $\delta_{M} (i; h)$:
$$\delta_{M} (i; h) = M(h, y) - M(h^{(-i)}, y^{(-i)})$$
If $\delta_{M} (i; h) > 0$, then $i$ is beneficial to $h$ with respect to $M$, as the inclusion of $h^{(i)}$ increases the corpus-level score. 
We define the \textit{favoritism} of $M$ toward $i$ as $\delta_{M} (i; h_S, h_U)$:
\begin{equation}
\delta_{M} (i; h_S, h_U) =\delta_{M} (i; h_S) - \delta_{M} (i; h_U)
\label{eq:deleterious}
\end{equation}
If $\delta_{M} (i; h_S, h_U) > 0$ then $M$ favors the translation of $x^{(i)}$ by system $S$ over that in system $U$. 

Table~\ref{tab:snmt_better_mf1} reflects the results of a manual examination of the ten sentences in the DE-EN test set with greatest magnitude favoritism; complete results are in the Appendix, Tables~\ref{tab:maf1-top-10} and \ref{tab:bleu-top-10}. Meaning-altering changes such as \textit{`untranslation'}, (wrong) \textit{`time'}, and (wrong) \textit{`translation'} are marked in \textit{italics}, while changes that do not fundamentally alter the meaning, such as `synonym,' (different) `inflection,' and (different) `word order' are marked in plain text.\footnote{Some changes, such as `word order' may change meaning; these are italicized or not on a case-by-case basis.}

\begin{table*}[htbp]
    \centering
    \resizebox{\textwidth}{!}{%
    \begin{tabular}{l|l}
$6^{th}$ & $\delta_\maf1(i,h_S,h_U)$: .044, $\delta_\bleu(i,h_S,h_U)$: -.00087, $\delta_{BLEURT}(i,h_S,h_U)$: .97\\\hline

Ref & Ever since I joined Labour 32 years ago as a school pupil, provoked by the Thatcher government's neglect \colorbox{yellow}{that had left my} \\
&\colorbox{yellow}{comprehensive school classroom literally falling down, I've sought to champion better public services for those who need them most} \\
&- \colorbox{yellow}{whether as a local councillor or government minister.}\\\hline

SNMT& 32 years ago, I joined Labour as a student because of the neglect of the Thatcher government, \colorbox{yellow}{which had led to my classroom literally} \\
&\colorbox{yellow}{collapsed, and as a result I tried to promote better public services for those who need it most, whether as a local council or ministers.}\\\hline

UNMT& Last 32 years ago, as a student, because of the disdain for the Thatcher-era government, Labour joined Labour.\\\hline

Problems & SNMT: synonym, word\_order	 UNMT: \textit{subject}, \colorbox{yellow}{\textit{truncation}}, \textit{word\_order}\\

    \end{tabular}
    }
    \caption{An example of favoritism that illustrates the differences between \maf1 and \bleu. Translations of the DE-EN test sentence with sixth largest magnitude favoritism according to \maf1, along with the favoritism according to \bleu\ (not in the top ten). UNMT's translation does not include the second half of the sentence. \maf1 favors SNMT, but \bleu\ favors UNMT.}
    \label{tab:diff_mf1_bleu_example}
\end{table*}

The results indicate that \maf1 generally favors SNMT, and with good reasons, as the favored translation does not generally alter sentence meaning, while the disfavored translation does. On the other hand, for the ten most favored sentences according to \bleu, four do not contain meaning-altering divergences in the disfavored translation. Importantly, none of the sentences with greatest favoritism according to \maf1, all of which having meaning altering changes in the disfavored alternatives, appears in the list for \bleu. This indicates relatively bad judgment on the part of \bleu. One case of good judgment from \maf1 and bad judgment from \bleu\ regarding truncation is shown in Table~\ref{tab:diff_mf1_bleu_example}. 

From our qualitative examinations, \maf1 is better than \bleu\ at discriminating against untranslations and trucations in UNMT. The case is similar for FR-EN and RO-EN, except that RO-EN has more untranslations for both SNMT and UNMT, possibly due to the smaller training data. Complete tables and annotated sentences are in the Appendix, in Section~\ref{app:extraqual}.

\section{Related Work}

\subsection{MT Metrics}
 Many metrics have been proposed for MT evaluation, which we broadly categorize into \textit{model-free} or \textit{model-based}. Model-free metrics compute scores based on translations but have no significant parameters or hyperparameters that must be tuned \textit{a priori}; these include  \bleu\ \cite{papineni-etal-2002-bleu}, NIST \cite{doddington2002-nist}, TER \cite{snover2006TER}, and \chrf1 \cite{popovic-2015-chrf}.  Model-based metrics have a significant number of parameters and, sometimes, external resources that must be set prior to use. These include METEOR \cite{banerjee-lavie-2005-meteor},  BLEURT \cite{sellam-etal-2020-bleurt}, YiSi \cite{lo-2019-yisi}, ESIM \cite{mathur-etal-2019-ESIM}, and BEER \cite{stanojevic-simaan-2014-beer}. Model-based metrics require significant effort and resources when adapting to a new language or domain, while model-free metrics require only a test set with references. 

\citet{mathur-etal-2020-tangled} have recently evaluated the utility of popular metrics and recommend the use of either \chrf1 or a model-based metric instead of \bleu. 
We compare our \maf1 and \mif1 metrics with \bleu, \chrf1, and BLEURT \cite{sellam-etal-2020-bleurt}. 
While \citet{mathur-etal-2020-tangled} use Pearson's correlation coefficient ($r$) to quantify the correlation between automatic evaluation metrics and human judgements, we instead use Kendall's rank coefficient ($\tau$), since $\tau$ is more robust to outliers than $r$ \cite{croux2010robust-correlation}. 

\subsection{Rare Words are Important}
\label{sec:rare-words}
That natural language word types roughly follow a Zipfian distribution is a well known phenomenon \cite{zipf1949human,powers-1998-zipf-apps}.
The frequent types are mainly so-called ``stop words,'' function words, and other low-information types, while most content words are infrequent types.
To counter this natural frequency-based imbalance, statistics such as inverted document frequency (IDF) are commonly used to weigh the \textit{input} words in applications such as information retrieval~\cite{Jones72specificity}.
In NLG tasks such as MT, where words are the \textit{output} of a classifier, there has been scant effort to address the imbalance.
\citet{doddington2002-nist} is the only work we know of in which the `information' of an n-gram is used as its weight, such that rare n-grams attain relatively more importance than in BLEU. 
We abandon this direction for two reasons:
Firstly, as noted in that work, \textit{large amounts of data are required to estimate n-gram statistics}.
Secondly, unequal weighing is a bias that is best suited to datasets where the weights are derived from, and such biases often do not generalize to other datasets.
Therefore, unlike \citet{doddington2002-nist}, we assign equal weights to all n-gram classes, and in this work we limit our scope to unigrams only.

While \bleu{} is a precision-oriented measure, METEOR \cite{banerjee-lavie-2005-meteor} and CHRF \cite{popovic-2015-chrf} include both precision and recall, similar to our methods.
However, neither of these measures try to address the natural imbalance of class distribution. 
BEER \cite{stanojevic-simaan-2014-beer} and METEOR \cite{denkowski-lavie-2011-meteor1.3} make an explicit distinction between function and content words; such a distinction inherently captures frequency differences since function words are often frequent and content words are often infrequent types. However, doing so requires the construction of potentially expensive linguistic resources. This work does not make any explicit  distinction and uses naturally occurring type counts to effect a similar result.

\subsection{F-measure as an Evaluation Metric}
F-measure \cite{Rijsbergen-1979-F-meas, chinchor-1992-F-meas} is extensively used as an evaluation metric in classification tasks such as part-of-speech tagging, named entity recognition, and sentiment analysis \cite{derczynski-2016-f-score}.
Viewing MT as a multi-class classifier is a relatively new paradigm \cite{gowda2020finding}, and evaluating MT solely as a multi-class classifier as proposed in this work is not an established practice.
However, we find that the $F_1$ measure is sometimes used for various analyses when \bleu{} and others are inadequate: The \texttt{compare-mt} tool \citep{neubig-etal-2019-compareMT} supports comparison of MT models based on $F_1$ measure of individual types.
\citet{gowda2020finding} use $F_1$ of individual types to uncover frequency-based bias in MT models.
\citet{sennrich-etal-2016-bpe} use corpus-level \textit{unigram $F_1$} in addition to \bleu\ and \chrf{}, however, corpus-level $F_1$ is computed as \mif1.
To the best of our knowledge, there is no previous work that clearly formulates the differences between micro- and macro- averages, and justifies the use of \maf1 for MT evaluation.

\section{Discussion and Conclusion}
We have evaluated NLG in general and MT specifically as a multi-class classifier, and illustrated the differences between micro- and macro- averages using \mif1 and \maf1 as examples (Section~\ref{sec:mt-as-cls}).
\maf1 captures semantic adequacy better than \mif1 (Section~\ref{sec:webnlg}).
\bleu, being a micro-averaged measure, served well in an era when generating fluent text was at least as difficult as generating adequate text. Since we are now in an era in which fluency is taken for granted and semantic adequacy is a key discriminating factor, macro-averaged measures such as \maf1 are better at judging the generation quality of MT models (Section~\ref{sec:wmt-metrics}).
We have found that another popular metric, \chrf1, also performs well on direct assessment,
however, being an implicitly micro-averaged measure, it does not perform as well as \maf1 on downstream CLIR tasks (Section~\ref{sec:material}).
Unlike BLEURT, which is also adequacy-oriented, \maf1 is directly interpretable, does not require retuning on expensive human evaluations when changing language or domain, and does not appear to have uncontrollable biases resulting from data effects.
It is both easy to understand and to calculate, and is  
inspectable, enabling fine-grained analysis at the level of individual word types. These attributes make it a useful metric for understanding and addressing the flaws of current models. For instance, we have used \maf1 to compare supervised and unsupervised NMT models at the same operating point measured in \bleu, and determined that supervised models have better adequacy than the current unsupervised models (Section~\ref{sec:unmt}).

Macro-average is a useful technique for addressing the importance of the long tail of language, and \maf1 is our first step in that direction; we anticipate the development of more advanced macro-averaged metrics that take advantage of higher-order and character n-grams in the future. 

\section{Ethical Consideration}

Since many ML models including NMT are themselves opaque and known to possess data-induced biases~\cite{prates2019-mt-bias}, using opaque and biased evaluation metrics in concurrence makes it even harder to discover and address the flaws in modeling.
Hence, we have raised concerns about the opaque nature of the current model-based evaluation metrics, and demonstrated examples displaying unwelcome biases in evaluation. We advocate the use of the \maf1 metric, as it is easily interpretable and offers the explanation of score as a composition of individual type performances.
In addition, \maf1 treats all types equally, and has no parameters that are directly or indirectly estimated from data sets. Unlike \maf1, \mif1 and other implicitly or explicitly micro-averaged metrics assign lower importance to rare concepts and their associated rare types. 
The use of micro-averaged metrics in real world evaluation could lead to marginalization of rare types.

\textit{Failure Modes:}
The proposed \maf1 metric is not the best measure of fluency of text. 
Hence we suggest caution while using \maf1 to draw fluency related decisions. \maf1 is inherently concerned with \textit{words}, and assumes the output language is easily segmentable into word tokens. Using \maf1 to evaluate translation into alphabetical languages such as Thai, Lao, and Khmer, that do not use white space to segment words, requires an effective tokenizer. Absent this the method may be ineffective; we have not tested it on languages beyond those listed in Section~\ref{sec:apphuman}.

\textit{Reproducibility:}
Our implementation of \maf1 and \mif1 has the same user experience as \bleu{} as implemented in \textsc{SacreBleu}; signatures are provided in Section~\ref{sec:appmetrics}. 
In addition, our implementation is  computationally efficient, and has the same (minimal) software and hardware requirements as \bleu{}. 
 All data for MT and NLG human correlation studies is publicly available and documented. Data for reproducing the IR experiments in Section~\ref{sec:lignos-etal} is also publicly available and documented. The data for reproducing the IR experiments in Section~\ref{sec:material} is only available to participants in the CLSSTS shared task. 

\textit{Climate Impact:} Our proposed metrics are on par with \bleu{} and such model-free methods, which consume significantly less energy than most model-based evaluation metrics.
\section*{Acknowledgements}

The authors thank Shantanu Agarwal, Joel Barry, and Scott Miller for their help with CLSSTS CLIR experiments, and Daniel Cohen for discussions on IR evaluation methods. This research is based upon work supported by the Office of the Director of National Intelligence (ODNI), Intelligence Advanced Research Projects Activity (IARPA), via AFRL Contract FA8650-17-C-9116.  The views and conclusions contained herein are those of the authors and should not be interpreted as necessarily representing the official policies or endorsements, either expressed or implied, of the ODNI, IARPA, or the U.S. Government. The U.S. Government is authorized to reproduce and distribute reprints for Governmental purposes notwithstanding any copyright annotation thereon.

\bibliography{80-refs}
\bibliographystyle{acl_natbib}

\clearpage
\appendix

\section{Metrics Reproducibility}
\label{sec:appmetrics}

\bleu\ scores reported in this work are computed with the \textsc{SacreBleu} library and have signature \texttt{\small BLEU+case.mixed+lang.<xx>-<yy>+numrefs.1 +smooth.exp+tok.<TOK>+version.1.4.13}, where \texttt{<TOK>} is \texttt{zh} for Chinese, and \texttt{13a} for all other languages. \maf1 and \mif1 use the same tokenizer as \bleu.
\chrf1 is also obtained using \textsc{SacreBleu} and has signature \texttt{\small chrF1+lang.<xx>-<yy>+numchars.6 +space.false +version.1.4.13}.
BLUERT scores are from the \textit{base} model of \citet{sellam-etal-2020-bleurt}, which is fine-tuned on WMT Metrics ratings data from 2015-2018.
The BLEURT model is retrieved from \url{https://storage.googleapis.com/bleurt-oss/bleurt-base-128.zip}.

\maf1 and \mif1 are computed using our fork of \textsc{SacreBleu} as:\\
\texttt{sacrebleu \$REF -m macrof microf < \$HYP}.

\section{Agreement with WMT Human Judgments}
\label{sec:apphuman}

Tables \ref{tab:wmt19-kendall}, \ref{tab:wmt18-kendall}, and \ref{tab:wmt17-kendall} provide $\tau$ between MT metrics and human judgments on WMT Metrics task 2017--2019. 
$\star$\bleu\ is based on pre-computed scores in WMT metrics package, whereas \bleu\ is based on our recalculation using \textsc{SacreBleu}. 
Values marked with \insig are not significant at $\alpha=0.05$, and hence corresponding rows are excluded from the calculation of mean, median, and standard deviation.

Since \maf1 is the only metric that does not achieve statistical significance in the WMT 2019 EN-ZH setting, we carefully inspected it.
Human scores for this setting are obtained without looking at the references by bilingual speakers \cite{WMT19-metrics-proceedings}, but the ZH references are found to have a large number of bracketed EN phrases, especially proper nouns that are rare types.
When the text inside these brackets is not generated by an MT system, \maf1 naturally penalizes heavily due to the poor recall.
Since other metrics assign lower importance to poor recall of such rare types, they achieve relatively better correlation to human scores than \maf1. 
However, since the $\tau$ values for EN-ZH are relatively lower than the other language pairs, we conclude that poor correlation of \maf1 in EN-ZH is due to poor quality references.
Some settings did not achieve statistical significance due to a smaller sample set as there were fewer MT systems submitted, e.g. 2017 CS-EN.

\begin{table}[ht]
    \footnotesize
    \centering
\begin{tabular}{l @{\hspace{1.5mm}} r @{\hspace{1.5mm}} r @{\hspace{1.5mm}} r @{\hspace{1.5mm}} r @{\hspace{1.5mm}} r}
 & $\star$\bleu & \bleu & \maf1 & \mif1 & \chrf1 \\ \hline \hline
DE-CS & .855 & .745 & .964 & .917 & \textbf{.982}  \\
DE-EN & .571 & .655 & .723 & .695 & \textbf{.742} \\
DE-FR & .782 & .881 & \textbf{.927} & .844 & .915 \\
EN-CS & .709 & \textbf{.954} & .927 & .927 & .908  \\
EN-DE & .540 & .752 & .741 & .773 & \textbf{.824} \\
EN-FI & .879 & .818 & .879 & .848 & \textbf{.923}  \\
EN-GU & .709 & .709 & .600 & \textbf{.734} & .709  \\
EN-KK & .491 & .527 & \textbf{.685} & .636 & .661  \\
EN-LT & .879 & .848 & \textbf{.970} & .939 & .881 \\
EN-RU & .870 & .848 & \textbf{.939} & .879 & .930  \\
FI-EN & .788 & .809 & \textbf{.909} & .901  & .875 \\
FR-DE & \textbf{.822} & .733 & .733 & .764  & .815 \\
GU-EN & .782 & .709 & .855 & .891 & \textbf{.945}  \\
KK-EN & \textbf{.891} & .844 & .796 & .844 & .881 \\
LT-EN & .818 & \textbf{.855} & .844 & \textbf{.855}  & .833 \\
RU-EN & .692 & .729 & .714 & \textbf{.780} & .757 \\
ZH-EN & .695 & .695 & \textbf{.752} & .676 & .715 \\ \hline
Median & .782 & .752 & .844 & .844 & .875\\
Mean & .751 & .771 & .821 & .818 & .841  \\
SD & .124 & .101 & .112 & .093 & .095  \\ \hline
EN-ZH & \textbf{.606} & \textbf{.606} & \insig.424 & .595 & .594 \\ \hline
Wins & 3 & 3 & 6 & 3 & 5 
\end{tabular} 
\caption{ WMT19 Metrics task: Kendall's $\tau$ between metrics and human judgments.}
\label{tab:wmt19-kendall}
\end{table}
%\vspace{2px}
\begin{table}[ht]
    \footnotesize
    \centering
\begin{tabular}{l @{\hspace{1.5mm}} r @{\hspace{1.5mm}} r @{\hspace{1.5mm}} r @{\hspace{1.5mm}} r @{\hspace{1.5mm}} r}
 & $\star$\bleu & \bleu & \maf1 & \mif1 & \chrf1 \\ \hline \hline
DE-EN & .828 & .845 & .917 & .883 & \textbf{.919}  \\
EN-DE & .778 & .750 & \textbf{.850} & .783 & .848  \\
EN-ET & .868 & .868 & .934 & .906 & \textbf{.949}  \\
EN-FI & .901 & .848 & .901 & .879 & \textbf{.945}  \\
EN-RU & .889 & .889 & \textbf{.944} & .889 & .930  \\
EN-ZH & .736 & .729 & .685 & \textbf{.833} & .827 \\
ET-EN & .884 & .900 & .884 & .878 & \textbf{.904}  \\
FI-EN & .944 & .944 & .889 & .915 & \textbf{.957}  \\
RU-EN & .786 & .786 & \textbf{.929} & .857 & .869 \\
ZH-EN & .824 & \textbf{.872} & .738 & .780 & .820  \\ 
EN-CS & \textbf{1.000} & \textbf{1.000} & .949 & \textbf{1.000} & .949  \\ \hline

Median & .868 & .868 & .901 & .879 & .919  \\
Mean   & .858 & .857 & .875 & .873 & .902  \\
SD     & .077 & .080 & .087 & .062 & .052  \\ \hline

TR-EN & \insig.200 & \insig.738 & \insig.400 & \insig.316 & \insig.632 \\
EN-TR & \insig.571 & \insig.400 & .837 & \insig.571 & \textbf{.849}  \\
CS-EN & \insig.800 & \insig.800 & \insig.600 & \insig.800 & \insig.738 \\ \hline
Wins &  1  &  2 & 3 &  2 & 6
\end{tabular}
\caption{ WMT18 Metrics task: Kendall's $\tau$ between metrics and human judgments.}
\label{tab:wmt18-kendall}

\end{table}
%\vspace{2px}
\begin{table}[ht]
    \footnotesize
    \centering
\begin{tabular}{l @{\hspace{1.5mm}} r @{\hspace{1.5mm}} r @{\hspace{1.5mm}} r @{\hspace{1.5mm}} r @{\hspace{1.5mm}} r}
 & $\star$\bleu & \bleu & \maf1 & \mif1 & \chrf1 \\ \hline \hline
DE-EN & .564 & .564 & .734 & .661 & \textbf{.744}  \\
EN-CS & .758 & .751 & .767 & .758 & \textbf{.878} \\
EN-DE & .714 & \textbf{.767} & .562 & .593 & .720  \\
EN-FI & .667 & .697 & .769 & .718 & \textbf{.782} \\
EN-RU & .556 & .556 & \textbf{.778} & .648 & .669  \\
EN-ZH & \textbf{.911} & \textbf{.911} & .600 & .854 & .899 \\
LV-EN & \textbf{.905} & .714 & \textbf{.905} & \textbf{.905} & \textbf{.905}  \\
RU-EN & .778 & .611 & .611 & .722 & \textbf{.800}  \\
TR-EN & \textbf{.911} & .778 & .674 & .733 & .907  \\
ZH-EN & .758 & \textbf{.780} & .736 & .824 & .732  \\ \hline
Median & .758 & .733 & .735 & .728 & .791 \\
Mean & .752 & .713 & .714 & .742 & .804  \\
SD & .132 & .110 & .103 & .097 & .088  \\ \hline
FI-EN & \textbf{.867} & \textbf{.867} & \insig.733 & \textbf{.867} & \textbf{.867} \\
EN-TR & \textbf{.857} & .714 & \insig.571 & .643 & .849 \\
CS-EN & \insig1.000 & \insig1.000 & \insig.667 & \insig.667 & \insig.913 \\  \hline
Wins & 5 & 4 & 2 & 2 & 6
\end{tabular} 
\caption{ WMT17 Metrics task: Kendall's $\tau$ between metrics and human judgments.}
\label{tab:wmt17-kendall}
\end{table}

\section{UNMT and SNMT Models}

\begin{table}[ht!]
% \resizebox{\textwidth}{!}{%
\centering
\footnotesize
\begin{tabular}{l @{\hspace{1.5mm}} r @{\hspace{1.5mm}} r @{\hspace{1.5mm}} l}
Translation & SNMT    & UNMT  &  SNMT Name      \\ \hline \hline
DE-EN NewsTest2019     & 32.7   & 33.9  & \textit{Our Transformer}             \\
EN-DE NewsTest2019     & 24.0   & 24.0  & \textit{Our Transformer}           \\
FR-EN NewsTest2014     & 31.1   & 31.2  & OnlineA.0             \\
EN-FR NewsTest2014     & 25.6   & 27.1  & \scriptsize{PROMT-Rule-based.3083} \\
RO-EN NewsTest2016     & 30.8   & 29.6  & Online-A.0            \\
EN-RO NewsTest2016     & 31.2   & 31.0  & uedin-pbmt.4362      \\ 
\end{tabular}%
% }
\caption{ SNMT systems are selected such that their \bleu\ scores are approximately the same as the available pretrained UNMT models.} 
\label{tab:unmt_vs_snmt2}
\end{table}

\begin{figure}[ht]
    \centering

    \begin{subfigure}[b]{0.9\linewidth}
    \includegraphics[width=\linewidth,trim={13mm 5mm 25mm 10mm},clip]{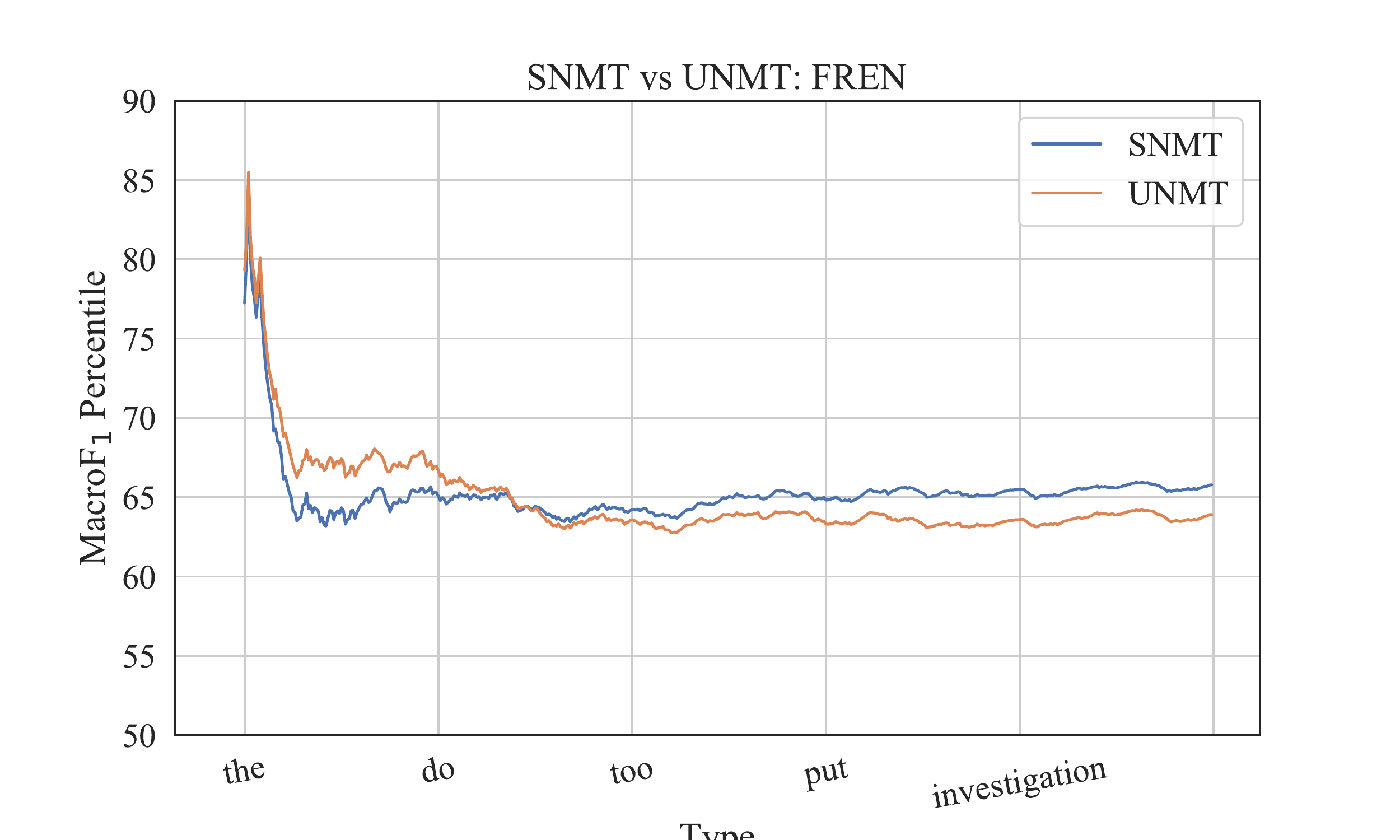}
    \end{subfigure}
    \hfill 
    \begin{subfigure}[b]{0.9\linewidth}
    \includegraphics[width=\linewidth,trim={13mm 7mm 25mm 10mm},clip]{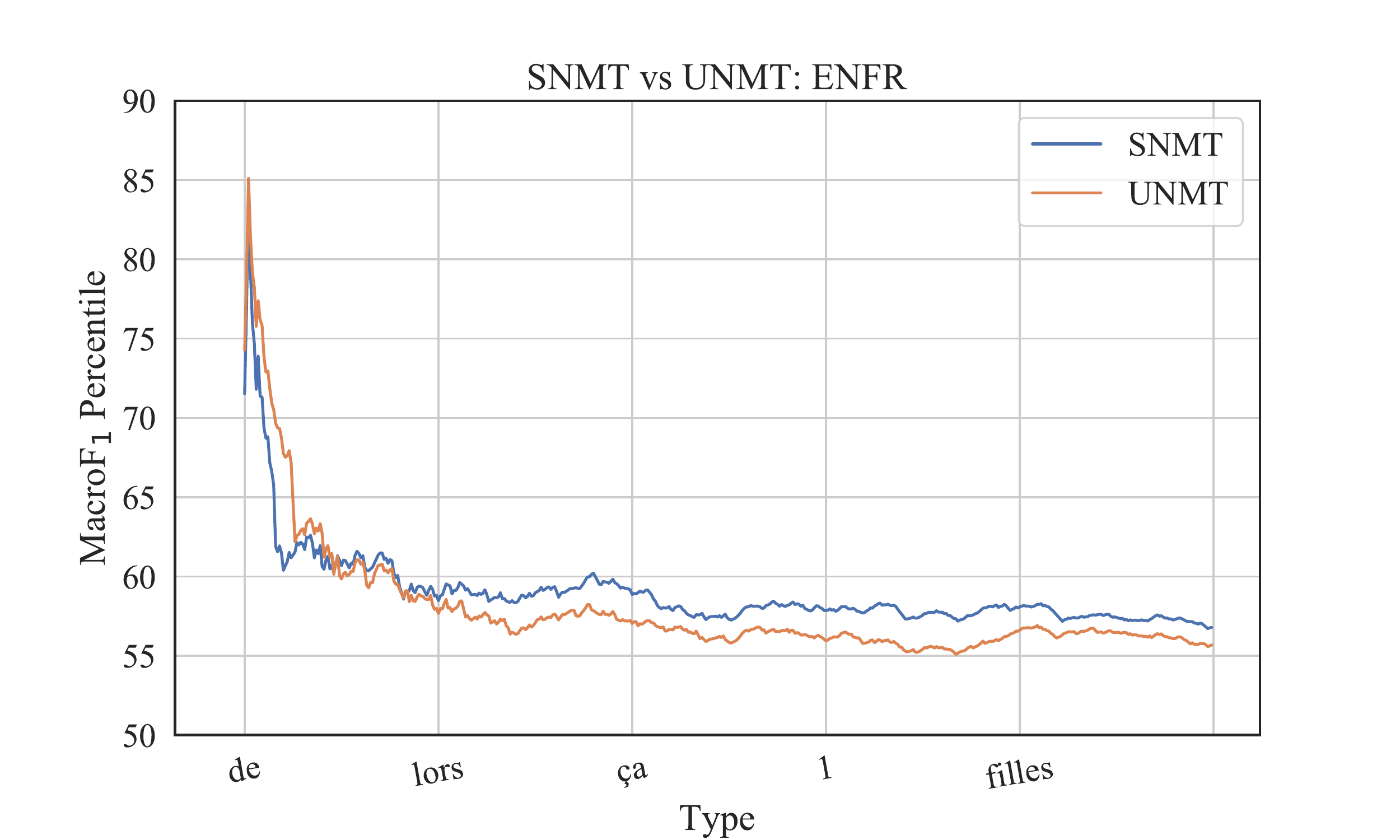}
    \end{subfigure}
    
    \begin{subfigure}[b]{0.9\linewidth}
    \includegraphics[width=\linewidth,trim={13mm 5mm 25mm 10mm},clip]{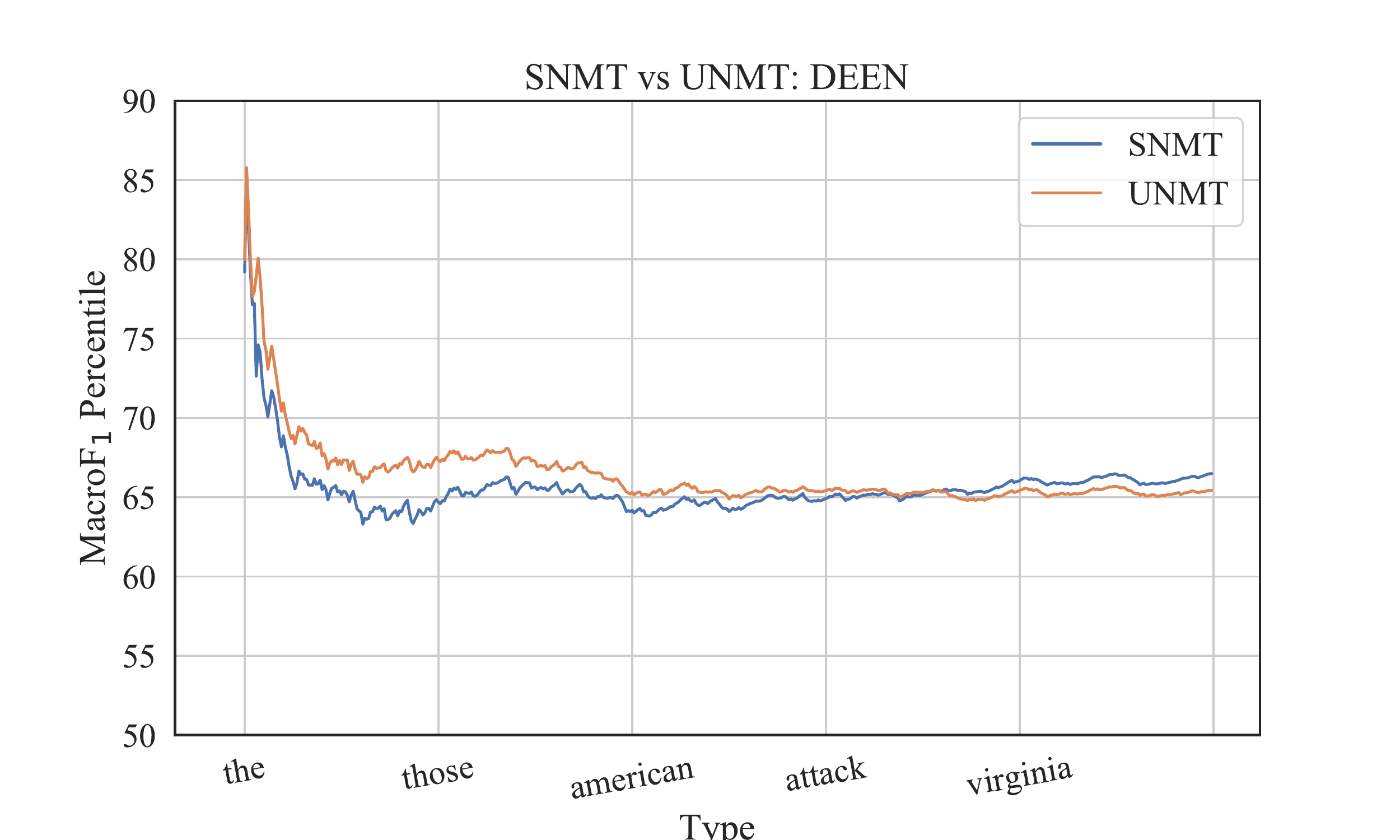}
    \end{subfigure}
    \hfill 
    \begin{subfigure}[b]{0.9\linewidth}
    \includegraphics[width=\linewidth,trim={13mm 7mm 25mm 10mm},clip]{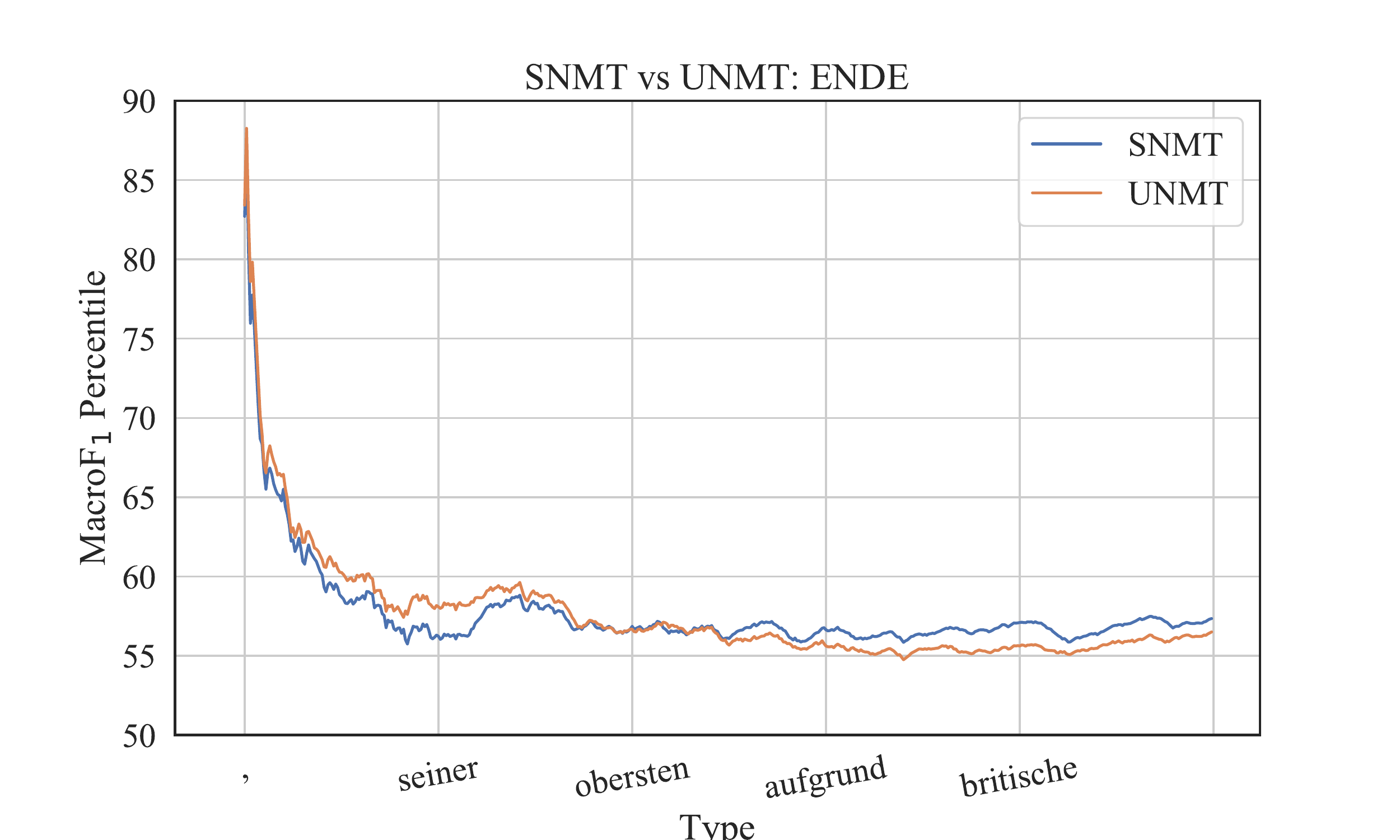}
    \end{subfigure}
    
    \begin{subfigure}[b]{0.9\linewidth}
    \includegraphics[width=\linewidth,trim={13mm 5mm 25mm 10mm},clip]{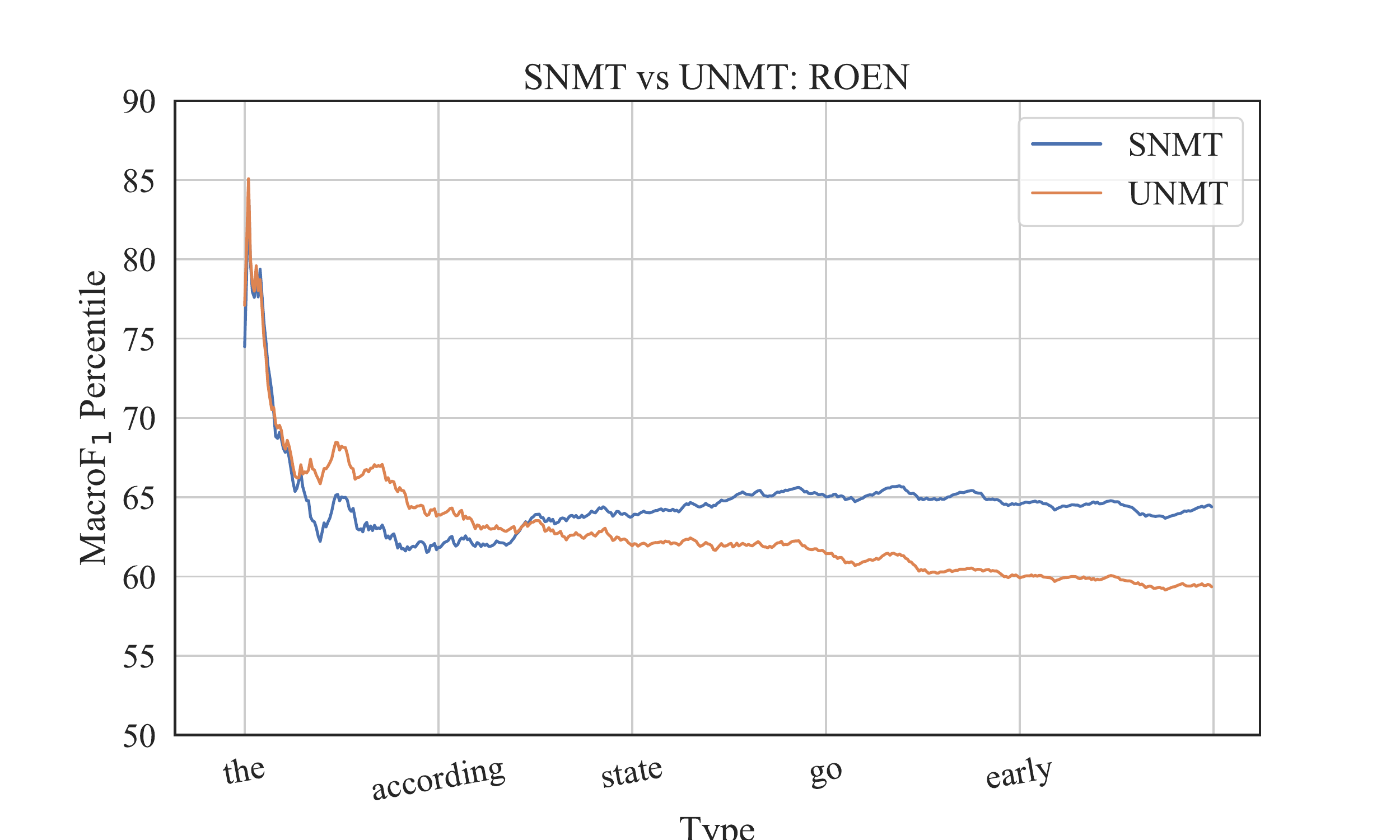}
    \end{subfigure}
\caption{SNMT vs UNMT \maf1 on the most frequent 500 types.
UNMT outperforms SNMT on frequent types that are weighed heavily by \bleu\, however, SNMT is generally better than UNMT on rare types; hence, SNMT has a higher \maf1. 
}
\label{fig:snmt_vs_unmt}
\end{figure}

% \begin{figure}[ht]
%     \centering
%     \includegraphics[width=\linewidth,trim={13mm 7mm 25mm 10mm},clip]{img/s_unmt-enro-maf1.pdf}
%     \label{fig:snmt_vs_unmt_rest}
%     \caption{(Continued from Figure~\ref{fig:snmt_vs_unmt}) Visualization of \maf1 between SNMT and UNMT .  }
% \end{figure}

The UNMT models follow XLM's standard architecture and are trained with 5 million monolingual sentences for each language using a vocabulary size of 60,000. 
% using code base from \citet{rtg-xt20}
We train SNMT models for EN$\leftrightarrow$DE and select models with the most similar (or a slightly lower) BLEU as their UNMT counterparts on newstest2019. The DE-EN model selected is trained with 1 million sentences of parallel data and a vocabulary size of 64,000, and the EN-DE model selected is trained with 250,000 sentences of parallel data and a vocabulary size of 48,000. For EN$\leftrightarrow$FR and EN$\leftrightarrow$RO, we select SNMT models from submitted systems to WMT shared tasks that have similar or slightly lower BLEU scores to corresponding UNMT models, based on \textit{NewsTest2014} for EN$\leftrightarrow$FR and \textit{NewsTest2016} for EN$\leftrightarrow$RO. 

Figure~\ref{fig:snmt_vs_unmt}, which is a visualization of \maf1 for SNMT and UNMT models,  shows that UNMT is generally better than SNMT on frequent types, however, SNMT outperforms UNMT on the rest leading to a crossover point in \maf1 curves. 
Since \maf1 assigns relatively higher weights to infrequent types than in \bleu{}, SNMT gains higher \maf1 than UNMT while both have approximately the same \bleu{}, as reported in Table~\ref{tab:unmt_vs_snmt}.

A complete comparison of UNMT vs SNMT in different languages is in Table~\ref{tab:unmt_vs_snmt2}. A manual analysis of the ten sentences with the largest magnitude favoritism according to \maf1 and \bleu\ in the FR-EN and RO-EN test sets is in Table~\ref{tab:snmt_better_mf1_fren} and Table~\ref{tab:snmt_better_mf1_roen}. The complete texts of these sentences, their reference translations, and the system translations (including DE-EN mentioned in Sec~\ref{sec:unmt}), are shown in Tables \ref{tab:maf1-top-10}, \ref{tab:bleu-top-10}, \ref{tab:maf1-top-10-fren}, \ref{tab:bleu-top-10-fren}, \ref{tab:maf1-top-10-roen}, and \ref{tab:bleu-top-10-roen}. 

\label{app:extraqual}

\begin{table*}[ht!]
    \centering
    \footnotesize
    \begin{tabular}{r @{\hspace{2mm}} l @{\hspace{2mm}} p{0.34\linewidth} | r @{\hspace{2mm}} l @{\hspace{2mm}} p{0.34\linewidth} }
 $\delta_{\maf1}$ & Fav & Analysis 
    & $\delta_{\bleu}$ & Fav   & Analysis \\ \hline \hline
 
 .044   & S  & S: synonym; U: \textit{untranslation}, synonym 
    &  -.026   & U  & U: synonym; S: \textit{omitted adv}, word order \\
 
 -.038   & U  & U: no issues; S: synonym 
    & .025   & S  & S: no issues; U: \textit{determiner}, word order \\ 
 
 .035  & S  & S: synonym; U: \textit{untranslation}, synonym    
    & .024   & S  & S: no issues; U: \textit{repetition}, form  \\

 -.034   & U  & U: no issues; S: synonym; word\_order
    & .021   & S  & S: \textit{verb}, synonym; U: \textit{untranslation}, \textit{noun}, \textit{time}, synonym \\ 
 
 -.034  & U  & U: synonym; S: \textit{word order}, verb\_ref  
  & .021  & S  & S: synonym; U: synonym  \\
 
 -.033   & U  & U: no issues; S: synonym
  &  -.021  & U  & U: \textit{omitted NER}; S: synonym, word order  \\
 
 .033   & S  & S: word order; U: \textit{untranslation}, NER, word order 
  & -.021  & U  & U: \textit{untranslation}; S: \textit{verb}, word order  \\
 
 .032  & S  & S: synonym; U: \textit{ number}, \textit{omitted noun}, \textit{untranslation}, \textit{verb} 
    & -.021  & U  & U: synonym; S: \textit{extra preposition}, synonym, word order \\
 
 .030  & S  &  S: \textit{adj}; U: \textit{untranslation}
  & .021  & S  & S: no issues; U: \textit{NER} \\
   
.030    & S & S: \textit{noun}, synonym; U: \textit{noun}, synonym  
  & -.020  & U  & U: synonym; S: synonym, word order 
\end{tabular}
\caption{Analysis of the ten FR-EN test set segments with the most favoritism in SNMT (S) or UNMT (U), according to $\maf1$ (left) and $\bleu$ (right). Fav is the favored system by metrics. Actual examples are shown in Appendix Tables~\ref{tab:maf1-top-10-fren} and \ref{tab:bleu-top-10-fren}.}
\label{tab:snmt_better_mf1_fren}
%\end{table*}

%\begin{table*}[ht!]
%    \centering
%    \footnotesize
    \begin{tabular}{r @{\hspace{2mm}} l @{\hspace{2mm}} p{0.34\linewidth} | r @{\hspace{2mm}} l @{\hspace{2mm}} p{0.34\linewidth} }
 $\delta_{\maf1}$ & Fav & Analysis 
    & $\delta_{\bleu}$ & Fav   & Analysis \\ \hline \hline
 
 .131   & S  & S: word order; U: \textit{repetition}, word order 
    &  .114   & S  & S: word order; U: \textit{repetition}, \textit{word order} \\
 
 .063   & S  & S: \textit{noun}, word order; U: \textit{repetition}, \textit{untranslation}, \textit{noun} 
    & .089   & S  & S: no issues; U: \textit{omitted noun}, \textit{omitted time}, \textit{NER} \\ 
 
 .062  & S  & S: \textit{extra}, \textit{untranslation}; U: \textit{untranslation}, \textit{copy}    
    & -.072   & U  & U: \textit{country}, \textit{untranslation}; S: \textit{noun}, \textit{word order}  \\

 -.052   & U  & U: \textit{untranslation x 3}, synonym; S: \textit{untranslation}, synonym
    & -.045   & U  & U: synonym; S: synonym, word order\\ 
 
 -.052  & U  & U: \textit{untranslation}, \textit{NER}, synonym; S: \textit{NER}, synonym 
  & -.041  & U  & U: \textit{untranslation}; S: \textit{word order}, \textit{subject}  \\
 
 -.052   & U  & U: \textit{extra}; S: \textit{untranslation}
  &  -.040  & U  & U: no issues; S: \textit{number}, \textit{omitted preposition}  \\
 
 -.050   & U  & U: \textit{adv}; S: \textit{incoherent}, \textit{adv}
  & .039  & S & S: \textit{extra}, \textit{untranslation}; U: \textit{untranslation}, \textit{copy}  \\
 
 -.050  & U  & U: \textit{active/passive voice}, \textit{name}; S: \textit{ name} 
    & .036  & S  & S: no issues; U: \textit{extra verb} \\
 
 -.049  & U  &  U: \textit{untranslation}; S: \textit{untranslation}, \textit{word order}
  & -.035  & U  & U: \textit{repetition}, \textit{untranslation}, S: \textit{verb}, synonym, word order \\
   
.048    & S & S: no issues; U: \textit{NER}
  & .034  & S  & S: synonym; U: \textit{untranslation}
\end{tabular}
\caption{Analysis of the ten RO-EN test set segments with the most favoritism in SNMT (S) or UNMT (U), according to $\maf1$ (left) and $\bleu$ (right). Fav is the favored system by metrics. Actual examples are shown in Appendix Tables~\ref{tab:maf1-top-10-roen} and \ref{tab:bleu-top-10-roen}.}
\label{tab:snmt_better_mf1_roen}
\end{table*}

\begin{table*}[ht]
%\scriptsize
\centering
\fontsize{8}{8}
\selectfont
%\scalebox{0.66}{
\begin{tabular}{r @{\hspace{2mm}} p{0.21\linewidth}p{0.2\linewidth}p{0.2\linewidth}p{0.2\linewidth}}
 $\delta_{\maf1}$&Source & Reference & SNMT & UNMT \\\hline\hline
.071 & Es wird davon ausgegangen, dass sie über eine leistungsstarke Kanone, eine Reihe von Flugabwehr- und Schiffsabwehrraketen sowie einige Stealth-Technologien verfügen, wie z. B. reduzierte Radar-, Infrarot- und akustische Signaturen.                                                                                                                                                         & It is understood they will feature a powerful cannon, an array of anti-aircraft and anti-ship missiles as well as some stealth technologies, such as reduced radar, infrared and acoustic signatures.                                                                                                                & It is assumed that they have a powerful cannon, a series of anti-aircraft and anti-ship missiles, as well as some steam technologies, such as reduced radar, infrared and acoustic signatures.                                                                                      & It is understood they have a powerful canon, a number of fluke and ship fire systems and some stealth-controlled technologies, such as reduced radar, infrarot and akustic signposts.                                                                                                    \\\hline
.064&Eine Gruppe maskierter Pro-Separatisten, die von der Bereitschaftspolizei zurückgehalten wurden, bewarfen sie mit Eiern und schleuderte Pulverfarbe und erzeugte in den Straßen, die normalerweise von Touristen überfüllt waren, dunkle Staubwolken.                                                                                                                                           & A group of masked pro-separatists held back by riot police pelted them with eggs and hurled powder paint, creating dark clouds of dust in streets that would usually be thronged with tourists.                                                                                                                      & A group of masked pro-separatists held hostage by the riot police brought them to eggs and ignited powder paint and produced dark clouds in the streets that were usually crowded by tourists.                                                                                      & A group of masked pro-independence separforces, who were kept away by the Bereitschaftpolice, beheaded them with evocative paint and poured pulverforce paint and created dark Staubations in the streets normally clogged by tourists.                                                  \\\hline
-.055&Il faut bien le faire.                                                                                                                                                                                                                                                                                                                                                                          & Il faut bien le faire.                                                                                                                                                                                                                                                                                               & La gentillesse du personnel et la disponibil.                                                                                                                                                                                                                                       & Il faut bien faire.                                                                                                                                                                                                                                                                      \\\hline
.052&In einem Abschnitt gibt es ein Bild eines Schlafsaals, in dem die Studenten auf Kaffeetassen, Vorhänge, Trainer und Bücher klicken, um über die Auswirkungen von Koffein und Licht informiert zu werden und darüber, wie sich die sportliche Leistung durch Schlafmangel und die Bedeutung einer Schlafenszeitroutine beeinflusst.                                                              & In one section there is an image of a dorm room, where students click on coffee cups, curtains, trainers and books to be told about the effects of caffeine and light and how athletic performance is impacted by sleep deficiency, and the importance of a bedtime routine.                                         & In a section there is an image of a bedroom where students click on coffee cups, curtains, trainers, and books to be informed about the effects of caffeine and light, and about how sporting performance is affected by lack of sleep and the importance of sleeping time routine. & In one section, there is a picture of a sleeping sauna where students click on coffee cups, forecourts, coaches and books to be educated about the impact of Koffein and light and about how athletic performance is influenced by sleep loss and the importance of a sleep day routine. \\\hline
-.045&Nickelbergbau ist auch in der Provinz wichtig, wird aber hauptsächlich in Morowali betrieben, an der gegenüberliegenden Küste von Sulawesi.                                                                                                                                                                                                                                                     & Nickel mining is also important in the province, but is mostly concentrated in Morowali, on the opposite coast of Sulawesi.                                                                                                                                                                                          & Nickelergbau is also important in the province, but is mainly operated in Morewali, on the opposite coast of Sulawesi.                                                                                                                                                              & Nickel mining is also important in the province, but is mostly operated in Morowali, on the opposite coast of Sulawesi.                                                                                                                                                                  \\\hline
.044&Vor 32 Jahren schloss ich mich als Schüler, wegen der Vernachlässigung der Thatcher-Regierung, Labour an. Diese Vernachlässigung hatte dazu geführt, dass mein Klassenzimmer buchstäblich zusammengebrochen war. Infolgedessen habe ich versucht, mich für bessere öffentliche Dienstleistungen für diejenigen einzusetzen, die sie am meisten brauchen. Egal ob als Gemeinderat oder Minister. & Ever since I joined Labour 32 years ago as a school pupil, provoked by the Thatcher government's neglect that had left my comprehensive school classroom literally falling down, I've sought to champion better public services for those who need them most - whether as a local councillor or government minister. & 32 years ago, I joined Labour as a student because of the neglect of the Thatcher government, which had led to my classroom literally collapsed, and as a result I tried to promote better public services for those who need it most, whether as a local council or ministers.     & Last 32 years ago, as a student, because of the disdain for the Thatcher-era government, Labour joined Labour.                                                                                                                                                                           \\\hline
.044&UN-Gesandter Staffan de Mistura hofft, bald die ersten Treffen eines neuen Ausschusses aus Regierungs- und Oppositionsmitgliedern einzuberufen, um eine Nachkriegsverfassung für Syrien zu entwerfen und den Weg zu Wahlen zu ebnen.                                                                                                                                                            & UN envoy Staffan de Mistura is hoping to soon convene the first meetings of a new committee comprised of government and opposition members to draft a post-war constitution for Syria and pave the way to elections.                                                                                                 & UN envoy Staffan de Mistura hopes to convene soon the first meetings of a new committee of government and opposition members to draw up a post-war constitution for Syria and pave the way for elections.                                                                           & U.N. Secretary General Staffan de Mistura hopes to soon join the first meetings of a new committee of government and opposition leaders to design a Nachkriegsrewrite for Syria and clear the path to elections.                                                                         \\\hline
.043&CBS hatte 3,1 Millionen, NBC 2,94 Millionen, MSNBC 2,89 Millionen und CNN 2,52 Millionen, so Nielsen.                                                                                                                                                                                                                                                                                           & CBS had 3.1 million, NBC had 2.94 million, MSNBC had 2.89 million and CNN had 2.52 million, Nielsen said.                                                                                                                                                                                                            & CBS had 3.1 million, NBC 2.94 million, MSNBC 2.89 million and CNN 2.52 million, says Nielsen.                                                                                                                                                                                       & CBS had 3.8 million, NBC 3.94 million, MSNBC 3.89 million and CNN 3.52 million, Nielsen said.                                                                                                                                                                                            \\\hline
-.041&Den Rangers gelangen nur zwei Schüsse in der ersten Hälfte, aber der ehemalige Ibrox-Torhüter Liam Kelly war kaum von Lassana Coulibalys Kopfsprung und dem Treffer eines bisslosen Ovie Ejaria aus der Ruhe zu bringen.                                                                                                                                                                        & Rangers managed just two first-half shots on target but former Ibrox goalkeeper Liam Kelly was barely troubled by Lassana Coulibaly's header and a tame Ovie Ejaria strike.                                                                                                                                          & The Ranners only reach two shots in the first half, but the former Ibrox-Torkeeper Liam Kelly was hardly the head of Lassanna Coulibys and the hit of a bissloze Ovi Ejaria.                                                                                                        & The Rangers managed only two shots in the first half but former Ibrox goalkeeper Liam Kelly was unlikely to be helped by Lassana Coulibaly's headfirst tackle and the goal of a bisected Ovie Ejaria.                                                                                    \\\hline
.041&Liverpool tritt am MIttwoch um 15.00 Uhr im Stadio San Paolo in Neapel, Italien, gegen Napoli an.                                                                                                                                                                                                                                                                                               & Liverpool battles Napoli in the group stage of the Champions League at 3 p.m. on Wednesday at Stadio San Paolo in Naples, Italy.                                                                                                                                                                                     & Liverpool will take place at 3 p.m. at the Stadio San Paolo in Naples, Italy, against Napoli.                                                                                                                                                                                       & Liverpool v Napoli at the MItch Stadium at 15.00 pm in Neapel, Italy, on MItch.
\end{tabular}
%}
\caption{Top 10 segments by $|\delta_{\maf1}(i, h_S, h_U)|$ on DE-EN.}
\label{tab:maf1-top-10}
\end{table*}

%%% extra extra stuff 
%\clearpage
%\input{91-appendix-tg}

\begin{table*}[ht]
%\scalebox{0.63}{
\centering
\fontsize{7.8}{7.8}
\selectfont
\begin{tabular}{r @{\hspace{1mm}} p{0.25\linewidth}p{0.2\linewidth}p{0.2\linewidth}p{0.2\linewidth}}
 $\delta_{\bleu}$ & Source   & Reference& SNMT & UNMT  \\\hline\hline
.048  & In der letzten Woche wurden mittlere Konzentrationen in Küstennähe und auf offener See in Pinellas County gemeldet, geringe bis hohe Konzentrationen auf offener See in Hillsborough County, Hintergrund- bis hohe Konzentrationen in Manatee County, Hintergrund- bis hohe Konzentrationen in Küstennähe und auf offener See in Sarasota County, Hintergrund- bis mittlere Konzentrationen in Charlotte County, Hintergrund- bis hohe Konzentrationen in Küstennähe und auf hoher See in Lee County sowie geringe Konzentrationen in Collier County. & Medium concentrations in or offshore of Pinellas County have been reported in the past week, low to high concentrations offshore of Hillsborough County, background to high concentrations in Manatee County, background to high concentrations in or offshore of Sarasota County, background to medium concentrations in Charlotte County, background to high concentrations in or offshore of Lee County, and low concentrations in Collier County. & Last week, average concentrations were reported on the coast and open seas in Pinellas County, low to high levels at open sea in Hillsborough County, background to high concentrations in Mantee County, high concentrations in coastal and open seas in Sarsota County, background to medium concentrations in Charlotte County, background to high shore and high sea levels in Lee County, and low concentrations in Collier County. & In the last week, moderate to high Konzentrof lead in Küstas County were reported in Pinellas County, low to high Konzentrof lead levels on open water in Hillsborough County, Hintergrundto high levels in Manatee County, Hintergrundto high to high Konzentrin Küstas and on open water in Sarasota County and low Konzentrationen in Charlotte County, Hintergrundto high to high Konzentrin Küstennähe and on open water in Sarasota County. \\\hline
.046  & Moskau hat wiederholt betont, dass die 11-Milliarden-Dollar-Pipeline Nord Stream 2, die die bestehende Pipeline-Kapazität auf 110 Milliarden Kubikmeter verdoppeln soll, ein rein wirtschaftliches Projekt ist.                                                                                                                                                                                                                                                                                                                                                               & Moscow has repeatedly stressed that the \$11 billion Nord Stream 2 pipeline, which is set to double the existing pipeline capacity to 110 billion cubic meters, is a purely economic project.                                                                                                                                                                                                                                                                                 & Moscow has repeatedly stressed that the \$11 billion Nord Stream 2 pipeline, which is supposed to double the existing pipeline capacity to 110 billion cubic metres, is a purely economic project.                                                                                                                                                                                                                                                               & Moscow has repeatedly insisted that the 11-billion pipeline, Nord Stream 2, which will double the existing Pipeline-capacity to 110 billion cubic feet, is a purely commercial project.                                                                                                                                                                                                                                                                                   \\
.044  & Der NTS, der für die Betreuung von mehr als 270 historischen Gebäuden, 38 wichtigen Gärten und 76.000 Hektar Land rund um das Land verantwortlich ist, nimmt die Fledermäuse sehr ernst.                                                                                                                                                                                                                                                                                                                                                                                      & The NTS, which is responsible for the care of more than 270 historical buildings, 38 important gardens and 76,000 hectares of land around the country, takes bats very seriously.                                                                                                                                                                                                                                                                                             & The NTS, which is responsible for the care of more than 270 historic buildings, 38 important gardens and 76,000 hectares of land around the country, takes the bats very seriously.                                                                                                                                                                                                                                                                              & The NTS, responsible for managing more than 270 historic buildings, 38 key gardens and 74,000 acres of land around the country, said the Fledermäuse are very important.                                                                                                                                                                                                                                                                                                  \\\hline
.042  & George W. Bush telefonierte mit Senatoren, um diese zu überreden, Herrn Kavanaugh zu unterstützen, der im Weißen Haus für Herrn Bush gearbeitet hatte und durch ihn seine Frau Ashley traf, die die persönliche Sekretärin von Herrn Bush war.                                                                                                                                                                                                                                                                                                                                & George W. Bush has been picking up the phone to call Senators, lobbying them to support Mr Kavanaugh, who worked in the White House for Mr Bush and through him met his wife Ashley, who was Mr Bush's personal secretary.                                                                                                                                                                                                                                                    & George W. Bush contacted senators to persuade them to support Mr Kavanaugh, who worked in the White House for Mr Bush and met his wife Ashley, who was Mr Bush's personal secretary.                                                                                                                                                                                                                                                                             & George W. Bush spoke to senators to help him overture to support Mr. Kavanaugh, who had worked in the White House for Mr. Bush and met through him his wife, Ashley, who was the personal secretary to Mr. Bush.                      \\\hline
-.039 & Eine Woche nachdem eine offizielle chinesische Zeitung eine vierseitige Anzeige in einer US-amerikanischen Tageszeitung auf den gegenseitigen Nutzen des US-China-Handels gestellt hatte, warf der US-amerikanische Botschafter in China Peking vor, die amerikanische Presse zur Verbreitung von Propaganda zu verwenden.                                                                                                                                                                                                                                                    & A week after an official Chinese newspaper ran a four-page ad in a U.S. daily touting the mutual benefits of U.S.-China trade, the U.S. ambassador to China accused Beijing of using the American press to spread propaganda.                                                                                                                                                                                                                                                 & A week after an official Chinese newspaper published a four-page display in a US daily on the mutual benefit of US China trade, the US ambassador to China published in Beijing to use the American press for propaganda.                                                                                                                                                                                                                                        & A week after an official Chinese newspaper published a four-page ad on the mutual benefit of the US-China trade, the U.S. ambassador to China accused Beijing of using the American press to spread propaganda.      \\\hline
-.037 & Sie kümmern sich nicht darum, wen sie verletzen, wen sie überfahren müssen, um Macht und Kontrolle zu bekommen, das ist, was sie wollen, Macht und Kontrolle, wir werden es ihnen nicht geben.                                                                                                                                                                                                                                                                                                                                                                                & They don't care who they hurt, who they have to run over in order to get power and control, that's what they want is power and control, we're not going to give it to them."                                                                                                                                                                                                                                                                                                  & They do not care about who they hurt whom they must pass over to gain power and control, that is what they want, power and control, we will not give them.                                                                                                                                                                                                                                                                                                       & They don't care who they hurt, who they have to pass to get power and control, that's what they want, power and control, we won't give it to them.                                                                                                                                                                                                                                                                                                                        \\\hline
-.034 & Mayorga behauptet, Ronaldo sei nach dem angeblichen Vorfall auf die Knie gefallen und habe ihr gesagt, er sei „zu 99 Prozent“ ein „guter Kerl“, der von den „ein Prozent“ im Stich gelassen wurde.                                                                                                                                                                                                                                                                                                                                                                            & Mayorga claims Ronaldo fell to his knees after the alleged incident and told her he was "99 percent" a "good guy" let down by the "one percent."                                                                                                                                                                                                                                                                                                                              & Mayorga claims that Ronaldo fell to the knees after the alleged incident, saying that he was “99\% ” a“ good guy ” left in the lurch by the “one percent ”.                                                                                                                                                                                                                                                                                                      & Mayorga claims Ronaldo fell on his knee after the alleged incident and told her he was "to 99 percent" a "good guy" who was left in the dark by the "one percent."                                                                                                                                                                                                                                                                                                        \\\hline
-.032 & Palin, 29, aus Wasilla, Alaska, wurde wegen des Verdachts auf häusliche Gewalt verhaftet. Gegen ihn liegt bereits ein Bericht über häusliche Gewalt und Widerstand bei der Festnahme vor, so eine Meldung, die am Samstag von den Alaska State Troopers veröffentlicht wurde.                                                                                                                                                                                                                                                                                                 & Palin, 29, of Wasilla, Alaska, was arrested on suspicion of domestic violence, interfering with a report of domestic violence and resisting arrest, according to a report released Saturday by Alaska State Troopers.                                                                                                                                                                                                                                                         & Palin, 29, from Wasilla, Alaska, was arrested for alleged domestic violence, and a report on domestic violence and opposition to arrest has already been published on Saturday by the Alaska State Trooperator.                                                                                                                                                                                                                                                  & Palin, 29, of Wasilla, Alaska, was arrested on charges of domestic violence. -- Against him, a report of domestic violence and resistance in arrest was already released Saturday, according to a report released Saturday by Alaska State Troopers.                                                                                                                                                                                                                      \\\hline
-.032 & "Ich habe {[}...{]} nicht versteckt Fords Behauptungen, ich habe ihre Geschichte nicht geleakt“, erzählte Feinstein dem Komitee, berichtete The Hill.                                                                                                                                                                                                                                                                                                                                                                                                                         & "I did not hide Dr. Ford's allegations, I did not leak her story," Feinstein told the committee, The Hill reported.                                                                                                                                                                                                                                                                                                                                                           & "I have {[}...{]} not hidden Ford's claims that I have not lived their history," told Finestein the committee, reported The Hill.                                                                                                                                                                                                                                                                                                                                & "I did not hide {[}Forman's claims, I didn't geleast her story," Feinstein told the committee, The Hill reported.                                                                                                                                                                                                                                                                                                                                                         \\\hline
-.031 & Briefings werden immer noch stattfinden, sagte Sanders, aber sollte die Presse die Chance haben, dem Präsidenten der Vereinigten Staaten die Fragen direkt zu stellen, so sei das unendlich besser, als mit ihr zu sprechen.                                                                                                                                                                                                                                                                                                                                                  & Briefings will still happen, Sanders said, but "if the press has the chance to ask the president of the United States questions directly, that's infinitely better than talking to me.                                                                                                                                                                                                                                                                                        & briefing is still going to take place, Sanders said, but the press should have the opportunity to put the questions directly to the President of the United States, if that is infinitely better than to talk to her.                                                                                                                                                                                                                                            & Briefings will still take place, Sanders said, but if the press has the chance to ask the president of the United States directly, so that is unendlich better than talking to her.                                                                                                                                                                                                                                                                                      
\end{tabular}
%}
\caption{Top 10 segments by $|\delta_{\bleu}(i, h_S, h_U)|$ on DE-EN.}
\label{tab:bleu-top-10}

\end{table*}

\begin{table*}[ht]
%\scriptsize
\centering
\fontsize{7.55}{7.55}
\selectfont
%\scalebox{0.66}{
\begin{tabular}{r @{\hspace{2mm}} p{0.22\linewidth}p{0.20\linewidth}p{0.20\linewidth}p{0.20\linewidth}}
 $\delta_{\maf1}$& Source & Reference & SNMT & UNMT \\\hline\hline
.044 & Il ne fallait qu'en déployer les accidents, et l'affaire, jacobinisme oblige, était confiée aux préfets et aux sous-préfets, interprètes autorisés. & All it took was to highlight its mistakes and, in keeping with Jacobinism, the issue would be entrusted to prefects and sub-prefects - the authorised interpreters. & It should deploy the accidents, and the case, Jacobinism obliges, was entrusted to the prefects and the sub-prefects, authorized interpreters. & It only took to deploy the accidents, and the matter, jacobinite oblige, was handed to the préfets and the sous-préfets, authorized interprètes. \\\hline
-.038 & Les spécialistes disent que les personnes sont systématiquement contraintes à faire leurs aveux, malgré un changement dans la loi qui a été voté plus tôt dans l'année interdisant aux autorités de forcer quiconque à s'incriminer lui-même. & Experts say confessions are still routinely coerced, despite a change in the law earlier this year banning the authorities from forcing anyone to incriminate themselves. & The experts say that people are systematically forced to make their confessions, despite a change in the law which was passed earlier this year prohibiting the authorities to force anyone to incriminating himself. & Experts say people are routinely forced to make their confessions, despite a change in the law that was passed earlier in the year banning officials from forcing anyone to incriminate themselves. \\\hline
.035 & Ils sont intersexués, l'intersexualité faisant partie du groupe de la soixantaine de maladies diagnostiquées comme désordres du développement sexuel, un terme générique désignant les personnes possédant des chromosomes ou des gonades (ovaires ou testicules) atypiques ou des organes sexuels anormalement développés. & They are intersex, part of a group of about 60 conditions that fall under the diagnosis of disorders of sexual development, an umbrella term for those with atypical chromosomes, gonads (ovaries or testes), or unusually developed genitalia. & They are intersex, intersex forming part of the Group of 60 diseases diagnosed as disorders of sexual development, a generic term for people with chromosomes or atypical gonads (ovaries or testes) or abnormally developed sexual organs. & They are intersexuzed, with intersexuality making up the group of the soixantaine of diseases diagnosed as disordered sexual development, a generic term dissignant people possessing chromosomes or gonades (ovaries or testicules) atypiques or anormally developed sexual organs. \\\hline
-.034 & Ces violences sont de plus en plus meurtrières en dépit de mesures de sécurité renforcées et d'opérations militaires d'envergure lancées depuis des mois par le gouvernement de Nouri Al Maliki, dominé par les chiites. & The violence is becoming more and more deadly in spite of reinforced security measures and large-scale military operations undertaken in recent months by Nouri Al Maliki's government, which is dominated by Shiites. & Such violence are more lethal despite measures enhanced security and large-scale military operations launched by the Government of Nouri Al Maliki, the Shia-dominated for months. & Those violence is increasingly deadly in the face of increased security measures and major military operations launched for months by Nouri Al Maliki's government, dominated by Shiites. \\\hline
-.034 & Du côté du gouvernement, on estime que 29 954 membres des forces armées du président Bachar el-Assad ont trouvé la mort, dont 18 678 étaient des combattants des forces pro-gouvernementales et 187 des militants du Hezbollah libanais. & On the government side, it said 29,954 are members of President Bashar Assad's armed forces, 18,678 are pro-government fighters and 187 are Lebanese Hezbollah militants. & On the side of the Government, it is estimated that 29 954 members of the armed forces of president Bachar Al-Assad died, whose 18 678 were 187 Lebanese Hezbollah militants and fighters of the pro-Government forces. & On the government side, one estimate says 29,954 members of President Bachar al-Assad's armed forces have found their way, including 18,678 were fighters from pro-government forces and 187 from Lebanese Hezbollah militants. \\\hline
-.033 & Mercredi, le Centre américain de contrôle et de prévention des maladies a publié une série de directives indiquant comment gérer les allergies alimentaires des enfants à l'école. & On Wednesday, the Centers for Disease Control and Prevention released a set of guidelines to manage children's food allergies at school. & Wednesday, the US Centre of disease prevention and control issued a set of guidelines indicating how to manage food allergies of children at the school. & Wednesday, the U.S. Centers for Disease Control and Prevention issued a series of directives indicating how to handle children's food allergies at school. \\\hline
.033 & N'est-il pas surprenant de lire dans les colonnes du Monde à quelques semaines d'intervalle d'une part la reproduction de la correspondance diplomatique américaine et d'autre part une condamnation des écoutes du Quai d'Orsay par la NSA ? & And is it not surprising to read in the pages of Le Monde, on the one hand, a reproduction of diplomatic correspondence with the US and, on the other, condemnation of the NSA's spying on the Ministry of Foreign Affairs on the Quai d'Orsay, within a matter of weeks? & Is it not surprising to read in the columns of the world a few weeks apart on the one hand the reproduction of American diplomatic correspondence and on the other hand a condemnation of the Quai d'Orsay by the NSA listens? & Isn't it surprising to read in the Times' pages just weeks apart of one side's reproduction of the American diplomatic correspondance and of another a condamnation of the Quai d' Orsay's écoutes by the NSA? \\\hline
.032 & Les ministères appellent à présent les personnes qui auraient été mordues, griffées, égratignées, ou léchées sur une muqueuse ou sur une peau lésée par ce chaton ou dont l'animal aurait été en contact avec ce chaton entre le 8 et le 28 octobre à contacter le 08.11.00.06.95 entre 10 heures et 18 heures à partir du 1er novembre. & The ministries are currently asking anyone who might have been bitten, clawed, scratched or licked on a mucous membrane or on damaged skin by the kitten, or who own an animal that may have been in contact with the kitten between 08 to 28 October, to contact them on 08 11 00 06 95 between 10am and 6pm from 01 November. & Departments now call people who have been bitten, scratched, scratched or licked on mucous membranes or skin injured by this kitten or where the animal would have been in contact with this kitten between 8 and 28 October to contact the 08.11.00.06.95 between 10 a.m. and 6 p.m. from November 1. & The at present call on people who may have been morbid, griffon, egregious or layed on a mug or on a skin léché by this chateau or whose animal may have been in contact with that chateau between 8 and 28 October to contact 08.11.00.095 between 10 and 18 November. \\\hline
.030 & A cette IIIe République, moment central et créateur, Pierre Nora a montré beaucoup d'intérêt et même de tendresse: saluant ceux qui se sont alors employés à réparer la fracture révolutionnaire, en enseignant aux écoliers tout ce qui dans l'ancienne France préparait obscurément la France moderne et en leur proposant une version unifiée de leur histoire. & Pierre Nora has shown has shown great interest and even tenderness for this Third Republic: he salutes those who tried at the time to repair the divide created by the Revolution by teaching students about everything in the former France that obscurely paved the way for the modern France, and by offering them a unified version of their history. & This third Republic, while central and creator, Pierre Nora has shown great interest and even tenderness: saluting those who then worked to repair the revolutionary divide, by teaching students what in the former France preparing darkly modern France and offering them a version unified in their history. & At this IIIe République, central and creator moment, Pierre Nora showed much interest and even tendresse: praising those who then helped to repair the revolutionary fracture, teaching schoolchildren everything in the former French Republic that obscurantly prepared modern France and offering them a unifying version of their history. \\\hline
.030 & La théorie dominante sur la façon de traiter les enfants pourvus d'organes sexuels ambigus a été lancée par le Dr John Money, de l'université Johns-Hopkins, qui considérait que le genre est malléable. & The prevailing theory on how to treat children with ambiguous genitalia was put forward by Dr. John Money at Johns Hopkins University, who held that gender was malleable. & The prevailing theory about how to treat children with ambiguous sexual organs was launched by Dr. John Money of the Johns Hopkins University, who considered that the genre is malleable. & The dominant theory about how to treat children armed with ambigüous sex organs was launched by Dr. John Money, of Johns-Hopkins University, who considered the genre maudlin. \\
\end{tabular}
%}
\caption{Top 10 segments by $|\delta_{\maf1}(i, h_S, h_U)|$ on FR-EN.}
\label{tab:maf1-top-10-fren}
\end{table*}

\begin{table*}[ht]
%\scriptsize
\centering
\fontsize{7.55}{7.55}
\selectfont
%\scalebox{0.66}{
\begin{tabular}{r @{\hspace{2mm}} p{0.22\linewidth}p{0.20\linewidth}p{0.20\linewidth}p{0.20\linewidth}}
 $\delta_{\bleu}$& Source & Reference & SNMT & UNMT \\\hline\hline
-.026 & Mais le représentant Bill Shuster (R-Pa.), président du Comité des transports de la Chambre des représentants, a déclaré qu'il le considérait aussi comme l'alternative la plus viable à long terme. & But Rep. Bill Shuster (R-Pa.), chairman of the House Transportation Committee, has said he, too, sees it as the most viable long-term alternative. & But Congressman Bill Shuster (R - PA.), Chairman of the House of representatives Transportation Committee, said that he considered as the most viable alternative in the long term. & But Rep. Bill Shuster (R-Pa.), chairman of the House Transportation Committee, said he also considered it the most viable long-term alternative. \\\hline
.025 & Les neuf premiers épisodes de Sheriff Callie's Wild West seront disponibles à partir du 24 novembre sur le site watchdisneyjunior.com ou via son application pour téléphones et tablettes. & The first nine episodes of Sheriff Callie's Wild West will be available from November 24 on the site watchdisneyjunior.com or via its application for mobile phones and tablets. & The first nine episodes of Sheriff Callie's Wild West will be available from November 24 on the site watchdisneyjunior.com or via its application for phones and tablets. & Sheriff first nine episodes of Sheriff Callie' s Wild West will be available as of November 24 on the watchdisneyjunior.com website or via its application for phones and computers. \\
.024 & Le président Xi Jinping, qui a pris ses fonctions en mars dernier, a fait de la lutte contre la corruption une priorité nationale, estimant que le phénomène constituait une menace à l'existence-même du Parti communiste. & President Xi Jinping, who took office last March, has made the fight against corruption a national priority, believing that the phenomenon is a threat to the very existence of the Communist Party. & President Xi Jinping, who took office in March, has made the fight against corruption a national priority, believing that the phenomenon posed a threat to the very existence of the Communist Party. & President Xi Jinping, who took office in March, has made fighting corruption a national priority, saying the phenomenon posed a threat to the Communist Party's existence-free existence. \\\hline
.021 & Un peu plus tôt, sur la route menant à Bunagana, poste-frontière avec l'Ouganda, des militaires aidés de civils chargeaient un lance-roquettes multiple monté sur un camion flambant neuf des FARDC, devant assurer la relève d'un autre engin pilonnant les positions du M23 sur les collines. & A little earlier, on the road to Bunagana, the frontier post with Uganda, soldiers assisted by civilians loaded up a multiple rocket launcher mounted on a brand new truck belonging to the FARDC, intended to take over from another device pounding the positions of the M23 in the hills. & Earlier, on the road leading to Bunagana border post with Uganda, soldiers helped civilians loaded a multiple rocket launcher mounted on a truck brand new FARDC, to ensure succession of another engine pounding the positions of the M23 in the hills. & A day earlier, on the road leading to Bunagana, a postcode with Uganda, military personnel aided by civilians were loading a multiple rocket lance-roquettes fire on a flambant neuf FARDC truck, expected to provide the lead for another device pilfering M23 positions on the hills. \\\hline
.021 & Il y a, avec la crémation, "une violence faite au corps aimé", qui va être "réduit à un tas de cendres" en très peu de temps, et non après un processus de décomposition, qui "accompagnerait les phases du deuil". & With cremation, there is a sense of "violence committed against the body of a loved one", which will be "reduced to a pile of ashes" in a very short time instead of after a process of decomposition that "would accompany the stages of grief". & There, with the cremation, "a violence made to the beloved body", which will be "reduced to a pile of ashes" in a very short time, and not after a process of decomposition, which "would accompany the phases of mourning". & There is, with cremation, "violence done to the loved one," which is going to be "reduced to a tas of ashes" in very little time, and not after a process of disablement, which would "accompany the phases of grief." \\\hline
.021 & Scott Brown, le capitaine du Celtic Glasgow, a vu son appel rejeté et sera bien suspendu pour les deux prochains matches de Ligue des champions de son club, contre l'Ajax et l'AC Milan. & Scott Brown, Glasgow Celtic captain, has had his appeal rejected and will miss his club's next two Champion's League matches, against Ajax and AC Milan. & Scott Brown, the captain of the Glasgow Celtic, saw his appeal dismissed and will be well suspended for the next two matches of the champions League for his club against Ajax and AC Milan. & Scott Brown, the Celtic captain, has had his appeal rejected and will be well suspended for his club's next two Champions League matches, against Ajax and AC Milan. \\\hline
-.021 & Les irréductibles du M23, soit quelques centaines de combattants, étaient retranchés à près de 2000 mètres d'altitude sur les collines agricoles de Chanzu, Runyonyi et Mbuzi, proches de Bunagana et Jomba, deux localités situées à environ 80 km au nord de Goma, la capitale de la province du Nord-Kivu. & The diehards of the M23, who are several hundreds in number, had entrenched themselves at an altitude of almost 2,000 metres in the farmland hills of Chanzu, Runyonyi and Mbuzi, close to Bunagana and Jomba, two towns located around 80km north of Goma, the capital of North Kivu province. & The irreducible m23, or a few hundred fighters, were cut off to nearly 2000 metres above sea level on the hills agricultural Chanzu, Runyonyi and Mbuzi, near Bunagana and Jomba, located about 80 km north of Goma, the capital of the province of North Kivu. & The irréductibles M23, or some hundred fighters, were retranchés at nearly 2000 feet of altitude on the agricultural hills of Chanzu, Runyonyi and Mbuzi, close to Bunagana and Jomba, two towns located about 80 miles north of Goma, the capital of North Kivu province. \\\hline
-.021 & Il a indiqué que le nouveau tribunal des médias « sera toujours partial car il s'agit d'un prolongement du gouvernement » et que les restrictions relatives au contenu et à la publicité nuiraient à la place du Kenya dans l'économie mondiale. & He said the new media tribunal "will always be biased because it's an extension of the government," and that restrictions on content and advertising would damage Kenya's place in the global economy. & He said as the new media tribunal ' will be always partial because it is an extension of the Government "and content and advertising restrictions hurt instead of the Kenya into the world economy. & He said the new media tribunal "will always be partial because it is a extension of the government" and that restrictions relating to content and advertising would hurt Kenya's place in the global economy. \\\hline
.021 & Dans "Les Fous de Benghazi", il avait été le premier à révéler l'existence d'un centre de commandement secret de la CIA dans cette ville, berceau de la révolte libyenne. & In "Les Fous de Benghazi", he was the first to reveal the existence of a secret CIA command centre in the city, the cradle of the Libyan revolt. & In "Les Fous de Benghazi", he was the first to reveal the existence of a secret CIA command center in this town, cradle of the Libyan revolt. & In "The Facts of Libya," he had been the first to reveal the existence of a secret CIA command center in that city, the birthplace of the Libyan uprising. \\\hline
-.020 & Le Sénat américain a approuvé un projet pilote de 90 M\$ l'année dernière qui aurait porté sur environ 10 000 voitures. & The U.S. Senate approved a \$90-million pilot project last year that would have involved about 10,000 cars. & The US Senate has approved a pilot project of 90 M\$ last year which would have covered about 10,000 cars. & The U.S. Senate approved a \$90 million pilot project last year that would have focused on about 10,000 cars. \\\hline
\end{tabular}
%}
\caption{Top 10 segments by $|\delta_{\bleu}(i, h_S, h_U)|$ on FR-EN.}
\label{tab:bleu-top-10-fren}
\end{table*}

\begin{table*}[ht]
%\scriptsize
\centering
\fontsize{7}{7}
\selectfont
%\scalebox{0.66}{
\begin{tabular}{r @{\hspace{1mm}} p{0.21\linewidth}p{0.21\linewidth}p{0.21\linewidth}p{0.24\linewidth}}
 $\delta_{\maf1}$& Source & Reference & SNMT & UNMT \\\hline\hline
.131 & David Grimal are o cariera internationala de violonist solo, de-a lungul careia a sustinut regulat concerte în ultimii 20 de ani pe principalele scene de muzica clasica ale lumii și cu orchestre prestigioase cum ar fi Orchestre de Paris, Orchestre Philharmonique de Radio France, Russian National Orchestra, Orchestre National de Lyon, Chamber Orchestra of Europe, Berliner Symphoniker, New Japan Philharmonic, Orchestre de l'Opera de Lyon, Mozarteum Orchestra Salzburg, Jerusalem Symphony Orchestra și Sinfonia Varsovia, sub conducerea unor dirijori precum Christoph Eschenbach, Michel Plasson, Michael Schnwandt, Peter Csaba, Heinrich Schiff, Lawrence Foster, Emmanuel Krivine, Mikhail Pletnev, Rafael Fruhbeck de Burgos și Peter Eotvos, Andris Nelsons, Christian Arming. & David Grimal has an international career as a solo violinist. During the last 20 years he held regular concerts on the main classical music stages of the world with prestigious orchestras such as the Orchestre de Paris, Orchestre Philharmonique de Radio France, Russian National Orchestra, Orchestre National de Lyon, Chamber Orchestra of Europe, Berliner Symphoniker, New Japan Philharmonic, Orchestre de l'Opera de Lyon, Salzburg Mozarteum Orchestra, Jerusalem Symphony Orchestra and Sinfonia Varsovia under the direction of conductors such as Christoph Eschenbach, Michel Plasson, Michael Schnwandt, Peter Csaba, Heinrich Schiff, Lawrence Foster, Emmanuel Krivine, Mikhail Pletnev, Rafael Fruhbeck de Burgos and Peter Eotvos, Andris Nelsons, Christian Arming. & David Grimal has an international career as a solo violinist, along which claimed concerts regularly in the past 20 years in key scenes of classical music of the world and with prestigious orchestras such as the Orchestre de Paris, Orchestre Philharmonique de Radio France, Russian National Orchestra, Orchestre National de Lyon, the Chamber Orchestra of Europe, the Berliner Symphoniker, the New Japan Philharmonic, Orchestre de l'Opera de Lyon The Salzburg Mozarteum Orchestra, the Jerusalem Symphony Orchestra, and Sinfonia Varsovia, under the direction of conductors such as Christoph Eschenbach, Michel Plasson, Michael Schnwandt, Peter Csaba, Heinrich Schiff, Lawrence Foster, Emmanuel Krivine, Mikhail Pletnev, Rafael Fruhbeck de Burgos, and Peter Eotvos, Andris Nelsons, Christian Arming. & David Grimal has an international solo violinist, Peter Csaba, Heinrich Schiff, Lawrence Foster, Emmanuel Krivine, Mikhail Pletnev, Mikhail Fruyhbeck Orchestra of Burgos and Peter Eotvos, Andris Philharmonique de Radio France, Russian National Orchestra, the French National Orchestra of Lyon, the Berliner Symphoniker, the New Japan Philharmonic, the Orchestre de l'Opera of Lyon, the Mozarteum Orchestra of Lyon, the Jerusalem Symphony Orchestra and Sinfonia of Paris, under the direction of conductor like Orchestra: the National Orchestra, the National Orchestra, the French Chamber Orchestra, the Berliner Symphoniology Orchestra and Sinfonia Sinfonia and Sinfonia Sinfonia Sinfonia, the Orchestra of the Paris Symphony Orchestra, the Russian National Orchestra, the National Orchestra, the National Opera of Europe, the Berliner Symphoniker, the Berliner Symphoniker Orchestra, the Orchestra of Lyon, the Mozarteum and Sinfonia Sinfonia Sinfonia. \\\hline
.063 & O parte importantă a magazinului este dedicată raionului uriaș de congelatoare, unde veți găsi frunze de curry, pepene și ghimbir congelate, rațe întregi, pește, sânge și bilă de vită, carcase de porc, chiftele de pește, cârnați tradiționali, semi-preparate și multe altele. & A good portion of the store is given over to the massive freezer section, where you'll find frozen curry leaves, bitter melon and galangal, whole ducks, fish, beef blood and bile, pork casings, fish balls, regional sausages, commercially-prepared foods and more. & An important part of the store is devoted to giant rayon of freezers, where you'll find curry leaves, melon and Ginger frozen whole fish, ducks, beef bile and blood, carcasses of pork, fish balls, sausages, traditional dishes and more. & A significant part of the store is devoted to the huge frozen fried chicken county, where you will find curry leaves, peach and frozen ghrelish, whole fish, fish, blood and bone stock, pork carcass, traditional fish carnouts, semi-cooked carnacies and many others. \\\hline
.062 & Persoanele interesate vor putea să afle cum să realizeze creații sculpturale florale armonioase (ikebana), cum să surprindă sufletul elementelor înconjurătoare în compoziții plastice folosind arta japoneză a pictării în tuș (Sumie) sau cum să își exprime propria individualitate și creativitate prin universul plurivalent al artelor frumoase (grafic, desen, pictură) , a declarat Sorin Mazilu, profesorul acestor cursuri. & Those interested will find out how to make harmonious sculptural floral creations (ikebana), how to capture the soul of the surrounding elements in plastic compositions using the Japanese ink painting art (Sumie) or how to express their own individuality and creativity through the multifaceted universe of Fine Arts (graphic, drawing, painting) said Sorin Mazilu, the teacher of these courses. & Persons interested will be able to learn how to achieve harmonious sculptural floral design (ikebana), how to capture the soul of the surrounding elements in plastic compositions using the Japanese art of painting in India ink (Sumie) or how to express their own individuality and creativity through the universe plurivalent of fine arts (painting, drawing, graphics), said Sorin Mazilu, teacher of such courses. & People interested will be able to learn how to complete armonious creative contrasting (grafic, drawing, painting) Creations, how to capture the hearts of surrounding elements in artistic compositions using the Japanese art of painting in tus (Sumie) or how to express their own individualistic identity through the plurivalent universe of beautiful arts (grafic, drawing, painting, painting), said Mr. Mazilu, the professor of these courses. \\\hline
-.052 & Umorul îmbrățișărilor frecvente ale lui Joey și Chandler, momentele în care se uită la fotbal în fotoliile comode și pasiunea lui Ross pentru Rachel au venit din faptul că se știe că bărbații se pot asocia cu această situație, chiar dacă adesea se feresc să își exploreze complet sentimentele. & The humour of Joey and Chandler's frequent hugs, moments watching football on the comfy chairs, and Ross's pining for Rachel, came from the knowledge that yes, men can all relate to this, even if they often hold back from fully exploring their feelings. & His humor frequently asked îmbrățișărilor of Joey and Chandler, the times when looking at football in comfortable armchairs and Ross's passion for Rachel came from the fact that it is known that men may be associated with this case, even if often keep out to explore fully the sentiments. & Ummaline frequent imbratisations of Joey and Chandler, the moments when he looks at football in his comfy fotoland Ross's passion for Rachel came from knowing that men can assimilate with the situation, even though often they are shy about fully exploring their feelings. \\\hline
-.052 & În plus, zilele următoare vom face recepția lucrărilor de la pasarela care leagă UPU de clinicile de cardiologie, hepatologie și gastroenterologie pentru a asigura transportul în condiții decente a bolnavului din UPU în clinicile vecine, a mai explicat managerul unității medicale. & In addition, in the coming days we will accept the works from the bridge linking the ER to the cardiology, hepatology and gastroenterology clinics to ensure the proper transfer of patients from the ER to the neighbouring clinics explained the manager of the medical unit. & In addition, the coming days we will make the reception of the works from the footbridge linking the UPU of Cardiology clinics, Hepatology and gastroenterology to ensure decent conditions of transportation of the patient from the UPU in neighboring clinics, explained Manager medical unit. & In addition, next week we will make the receptia work on the pasarela linking the UPU to the cardiology, hepatology and gastroenterology clinicians to ensure the transport in decent manner of the patient from the UPU to neighbouring clinicians, the hospital's chief executive explained. \\\hline
-.052 & Desi presa araba anunta ca despartirea lui Sanmartean de Ittihad e iminenta, impresarul antreorului Ladislau Boloni, Arcadie Zaporojanu, spune ca acest lucru nu se va mai intampla. & Although Arab media announced that the separation between Sanmartean and Ittihad is imminent, coach Ladislau Boloni's agent, Arcadia Zaporojanu, says this will no longer happen. & Although the Arabic press announces that the breakup of his impending Sanmartean Ittihad e, impresario, Arcadie antreorului Baghery Isaac says that this will no longer happen. & While the Arab media say Sanmartean's departure from Ittihad is imminent, impressions of coach Ladislau Boloni's agent Arcadie Zaporojanu say this will never happen. \\\hline
-.05 & Cei mai putini sunt cei care au finalizat doctoratul - 0,67 - și tinerii care au terminat liceul ori scoala generala - 6,20\%. & The least numerous are those who completed the doctorate courses - 0.67 - and young people who have finished high school or middle school - 6.20\%. & Most are few people who have completed their Ph.d.-0.67-and young people who have completed grade school-high school times 6,20\%. & Most are those who completed their degree - 0.67 - and young people who finished college or high school - 6.20 percent. \\\hline
-.05 & Politica clubului a facut din Viitorul o echipa de urmarit pentru echipele din strainatate, astfel ca Hagi a vandut în aceasta vara de 1,5 milioane de euro - Ianis Hagi (Fiorentina - 1 milion de euro) și Alexandru Mitrita (Bari - 500 mii euro). & The club's policy made Viitorul a team to be watched by foreign teams, and so Hagi has sold this summer in the amount of 1.5 million Euros - Ianis Hagi (Fiorentina - 1 million euro) and Alexandru Mitrita (Bari - 500 thousand euro). & The Club's policy has made the team watched for foreign teams, such as Hagi has sold in this summer of 1.5 million euros-Ibarra H (Fiorentina-EUR 1 million) and Alexander Duane (Bari-500 thousand euros). & The club's policy made the Viitorul a team to watch for overseas teams, so Hagi sold him for £1.5 million - Ianis Hagi (Fiorentina - £1 million) and John Mitrita (Bari - £500 million). \\\hline
-.049 & Au inclus amestecul de pungași neplătitori de taxe, străini care fraudau telefonic și pseudo non-domi care dețin majoritatea ziarelor importante. & They included the medley of tax-shy rascals, phone-hacking foreigners and pseudo non-doms who own most of our great newspapers. & They included a mixture of tax revenue, pungași foreign fraudau telephone and non-pseudo domi who hold most major newspapers. & They included the mixture of unpaid tax pungas, foreigners who telephone and pseudo non-doms who own most of the key newspapers. \\\hline
.048 & Constantin Brâncuși se naște la 19 februarie 1876, în Hobița, un mic sat din comuna Peștișani, județul Gorj, la poalele Carpaților. & Constantin Brancusi was born on 19 February 1876 in Hobița, a small village from Peștișani, Gorj county, at the foot of the Carpathians. & Constantin Brâncuși February 19, was born in 1876, a small Certified village in Peștișani commune, Gorj County, at the foot of the Carpathians. & Mr. Brancusi was born Feb. 19, 1876, in Hobson, a small village in the town of Peston, Pa., at the foot of the mountains. \\\hline
\end{tabular}
%}
\caption{Top 10 segments by $|\delta_{\maf1}(i, h_S, h_U)|$ on RO-EN.}
\label{tab:maf1-top-10-roen}
\end{table*}

\begin{table*}[ht]
%\scriptsize
\centering
\fontsize{6.5}{6.5}
\selectfont
%\scalebox{0.66}{
\begin{tabular}{r @{\hspace{1mm}} p{0.21\linewidth}p{0.21\linewidth}p{0.21\linewidth}p{0.24\linewidth}}
 $\delta_{\bleu}$& Source & Reference & SNMT & UNMT \\\hline\hline
.114 & David Grimal are o cariera internationala de violonist solo, de-a lungul careia a sustinut regulat concerte în ultimii 20 de ani pe principalele scene de muzica clasica ale lumii și cu orchestre prestigioase cum ar fi Orchestre de Paris, Orchestre Philharmonique de Radio France, Russian National Orchestra, Orchestre National de Lyon, Chamber Orchestra of Europe, Berliner Symphoniker, New Japan Philharmonic, Orchestre de l'Opera de Lyon, Mozarteum Orchestra Salzburg, Jerusalem Symphony Orchestra și Sinfonia Varsovia, sub conducerea unor dirijori precum Christoph Eschenbach, Michel Plasson, Michael Schnwandt, Peter Csaba, Heinrich Schiff, Lawrence Foster, Emmanuel Krivine, Mikhail Pletnev, Rafael Fruhbeck de Burgos și Peter Eotvos, Andris Nelsons, Christian Arming. & David Grimal has an international career as a solo violinist. During the last 20 years he held regular concerts on the main classical music stages of the world with prestigious orchestras such as the Orchestre de Paris, Orchestre Philharmonique de Radio France, Russian National Orchestra, Orchestre National de Lyon, Chamber Orchestra of Europe, Berliner Symphoniker, New Japan Philharmonic, Orchestre de l'Opera de Lyon, Salzburg Mozarteum Orchestra, Jerusalem Symphony Orchestra and Sinfonia Varsovia under the direction of conductors such as Christoph Eschenbach, Michel Plasson, Michael Schnwandt, Peter Csaba, Heinrich Schiff, Lawrence Foster, Emmanuel Krivine, Mikhail Pletnev, Rafael Fruhbeck de Burgos and Peter Eotvos, Andris Nelsons, Christian Arming. & David Grimal has an international career as a solo violinist, along which claimed concerts regularly in the past 20 years in key scenes of classical music of the world and with prestigious orchestras such as the Orchestre de Paris, Orchestre Philharmonique de Radio France, Russian National Orchestra, Orchestre National de Lyon, the Chamber Orchestra of Europe, the Berliner Symphoniker, the New Japan Philharmonic, Orchestre de l'Opera de Lyon The Salzburg Mozarteum Orchestra, the Jerusalem Symphony Orchestra, and Sinfonia Varsovia, under the direction of conductors such as Christoph Eschenbach, Michel Plasson, Michael Schnwandt, Peter Csaba, Heinrich Schiff, Lawrence Foster, Emmanuel Krivine, Mikhail Pletnev, Rafael Fruhbeck de Burgos, and Peter Eotvos, Andris Nelsons, Christian Arming. & David Grimal has an international solo violinist, Peter Csaba, Heinrich Schiff, Lawrence Foster, Emmanuel Krivine, Mikhail Pletnev, Mikhail Fruyhbeck Orchestra of Burgos and Peter Eotvos, Andris Philharmonique de Radio France, Russian National Orchestra, the French National Orchestra of Lyon, the Berliner Symphoniker, the New Japan Philharmonic, the Orchestre de l'Opera of Lyon, the Mozarteum Orchestra of Lyon, the Jerusalem Symphony Orchestra and Sinfonia of Paris, under the direction of conductor like Orchestra: the National Orchestra, the National Orchestra, the French Chamber Orchestra, the Berliner Symphoniology Orchestra and Sinfonia Sinfonia and Sinfonia Sinfonia Sinfonia, the Orchestra of the Paris Symphony Orchestra, the Russian National Orchestra, the National Orchestra, the National Opera of Europe, the Berliner Symphoniker, the Berliner Symphoniker Orchestra, the Orchestra of Lyon, the Mozarteum and Sinfonia Sinfonia Sinfonia. \\\hline
.089 & Editia de toamna va avea loc la Chisinau (Casa Armatei, 2-3 octombrie), Iasi (Palas, 6-7 octombrie), Brasov (Casa Armatei, 14-15 octombrie), Cluj Global (Sala Polivalenta, 20-21 octombrie), Cluj IT (Sala Polivalenta, 22-23 octombrie), Sibiu (Centrul de Afaceri, 28-29 octombrie) și Targu-Mures (Teatrul National, 5 noiembrie). & The autumn edition will take place in Chisinau (Army House 2-3 October), Iasi (Palas 6-7 October), Brasov (Army House 14-15 October), Cluj Global (Polyvalent Hall 20-21 October) Cluj IT (Polyvalent Hall 22-23 October), Sibiu (Business Centre 28-29 October) and Targu-Mures (National Theatre, November 5). & Autumn edition will take place in Chisinau (Army House, 2-3 October), Iasi (Palas, 6-7 October), Brasov (Army House, 14-16 October), Global (Sala Polivalenta, 20-21 October), Cluj (Sala Polivalenta, 22-23 October), Sibiu (Business Centre, 28-29 October) and Targu-Mures (National Theatre, November 5). & The fall season will be in Washington (House, 2-3), Washington (Palas, Oct. 6-7), Brasov (House, Oct. 14-15), San Francisco Global (East Coast, Oct. 20-21), San Francisco IT (East Coast, Oct. 22-23), Sibiu (Center for Business, Oct. 28-29) and Targu-Mures (the National Center, Nov. 5). \\\hline
-.072 & Realizatorii studiului mai transmit ca "romanii simt nevoie de ceva mai multa aventura în viata lor (24\%), urmat de afectiune (21\%), bani (21\%), siguranta (20\%), nou (19\%), sex (19\%), respect 18\%, incredere 17\%, placere 17\%, conectare 17\%, cunoastere 16\%, protectie 14\%, importanta 14\%, invatare 12\%, libertate 11\%, autocunoastere 10\% și control 7\%". & The study's conductors transmit that "Romanians feel the need for a little more adventure in their lives (24\%), followed by affection (21\%), money (21\%), safety (20\%), new things (19\%), sex ( 19\%) respect 18\%, confidence 17\%, pleasure 17\%, connection 17\%, knowledge 16\%, protection 14\%, importance 14\%, learning 12\%, freedom 11\%, self-awareness 10\% and control 7\% ". & The filmmakers may study transmitted as "Romanians feel in need of something more adventure in their lives (24\%), followed by affection (21\%), money (21\%), security (20\%), new (19\%), sex (19\%), respect, trust and 18\% 17\% 17\% pleasure, 17\%, 16\%, knowledge protection 14\%, 14\%, 12\%, liberty learning 11\%, self-awareness and control 10\% 7\%". & Reporters also deliver that "Americans feel need for some more adventure in their lives (24\%), followed by affection (21\%), money (21\%), safety (20\%), new (19\%), sex (19\%), respect 18\%, confidence 17\%, pleasure 17\%, connection 17\%, knowledge 14\%, protection 14\%, importance 14\%, learning 12\%, freedom 11\%, autocunoastere 10\% and control 7\%." \\\hline
-.045 & Fed ar trebui să pună problema stabilității financiare pe primul loc doar în cazul unei crize majore, cum a fost cutremurul de pe piață din 2008, a declarat Adam S. Posen, fost membru al comisiei de stabilire a ratei dobânzii din cadrul Bank of England. & The Fed should put financial stability concerns first only during a major crisis, such as the 2008 market meltdown, said Adam S. Posen, a former member of the Bank of England's rate-setting committee. & The Fed should put the issue of financial stability in the first place only in the event of major crises such as the earthquake in the market since 2008, "said Adam s. Posen, a former member of the Commission for determining the interest rate within the Bank of England. & The Fed should put financial stability first only in a major crisis, such as the 2008 credit crunch, said Adam S. Posen, a former member of the Bank of England's stabilisation committee. \\\hline
-.041 & Atunci când susținătorilor lui Clinton li se adresează o întrebare deschisă referitor la motivul pentru care ar dori ca ea să câștige cursa, răspunsul majoritar este că deține experiența adecvată (16\%), urmat de faptul că a venit momentul ca o femeie să fie președinte (13\%) și că este cel mai bun candidat pentru această funcție (10\%). & When Clinton's supporters are asked in an open-ended question why they want her to be the nominee, the top answer is that she has the right experience (16 percent), followed by it's time for a woman president (13 percent), and that she is the best candidate for the job (10 percent). & When Clinton supporters are asked to answer an open question regarding why you want her to win the race, the answer is that majority holds appropriate experience (16\%), followed by the fact that the time has come for a woman to be President (13\%), and that is the best candidate for this position (10\%). & When Clinton supporters are asking an open question as to why she would like to win the race, the majoritar answer is that she holds the right experience (16 percent), followed by the fact that it has come time for a woman to be president (13 percent) and that she is the best candidate for the job (10 percent). \\\hline
-.040 & Clinton este acum susținută de 47\% din alegătorii democrați (în scădere de la 58\%), în timp ce Sanders este pe locul doi, cu 27\% (în urcare față de 17\%). & Clinton now has the backing of 47 percent of Democratic primary voters (down from 58 percent), while Sanders comes in second, with 27 percent (up from 17 percent). & Clinton is now supported by 47\% of the voters Democrats (down 62\%), while Sanders is in second place with 27\% (in relation to climb 17\%). & Clinton is now backed by 47 percent of Democratic voters (down from 58 percent), while Sanders is second with 27 percent (up from 17 percent). \\\hline
.039 & Persoanele interesate vor putea să afle cum să realizeze creații sculpturale florale armonioase (ikebana), cum să surprindă sufletul elementelor înconjurătoare în compoziții plastice folosind arta japoneză a pictării în tuș (Sumie) sau cum să își exprime propria individualitate și creativitate prin universul plurivalent al artelor frumoase (grafic, desen, pictură) , a declarat Sorin Mazilu, profesorul acestor cursuri. & Those interested will find out how to make harmonious sculptural floral creations (ikebana), how to capture the soul of the surrounding elements in plastic compositions using the Japanese ink painting art (Sumie) or how to express their own individuality and creativity through the multifaceted universe of Fine Arts (graphic, drawing, painting) said Sorin Mazilu, the teacher of these courses. & Persons interested will be able to learn how to achieve harmonious sculptural floral design (ikebana), how to capture the soul of the surrounding elements in plastic compositions using the Japanese art of painting in India ink (Sumie) or how to express their own individuality and creativity through the universe plurivalent of fine arts (painting, drawing, graphics), said Sorin Mazilu, teacher of such courses. & People interested will be able to learn how to complete armonious creative contrasting (grafic, drawing, painting) Creations, how to capture the hearts of surrounding elements in artistic compositions using the Japanese art of painting in tus (Sumie) or how to express their own individualistic identity through the plurivalent universe of beautiful arts (grafic, drawing, painting, painting), said Mr. Mazilu, the professor of these courses. \\\hline
.036 & Companiile au vacante cateva mii de locuri de munca, oportunitati de internship și stagii de practica, iar o parte dintre ele sunt deja anuntate pe site-ul oficial www.targuldecariere.ro. & Companies have thousands of vacant jobs, internship opportunities and internships, and some of them are already posted on the official website www.targuldecariere.ro. & Companies have thousands of vacant jobs, internship opportunities, and internships, and some of them are already announced on the official website www.targuldecariere.ro. & Companies have filled several thousand jobs, internships and training trips, and a number of them are already announced on the official website www.targuldecarier.com. \\\hline
-.035 & Antrenorul celor de la Brisbane Broncos, Wayne Bennett, a făcut o ușoară referire la Storm după victoria echipei sale în meciul de calificare cu North Queensland Cowboys, jucat sâmbătă seară, când a numit meciul respectiv o „demonstrație” a ligii de rugby și a declarat că cele două echipe din Queensland nu se prea pricep la wrestling. & Brisbane Broncos coach Wayne Bennett made a thinly veiled reference to the Storm after his side's qualifying final win over North Queensland Cowboys on Saturday night when he called that game a "showcase" of the rugby league and said the two Queensland weren't "too big" into wrestling. & The coach of the Brisbane Broncos, Wayne Bennett, made a slight reference to the Storm after his team's victory in the qualifying match with North Queensland Cowboys, played Saturday night, when he called the match a "demonstration" of rugby league and said that the two teams from Queensland is not too good at wrestling. & Brisbane Broncos coach Wayne Bennett made a slight reference to the Storm after their team's victory in the qualifying match against the North Queensland Cowboys on Saturday night when he called the game a "demonstration" of the rugby league league league and said the two Queensland sides were not really pricep to wrestling. \\\hline
.034 & David Grimal recunoaste ca Les Dissonaces reprezinta "un lux", precizand ca niciunul dintre muzicienii care fac parte din orchestra nu depinde de succesul acestui proiect pentru a trai. & David Grimal admits that Les Dissonaces is "a luxury", adding that none of the musicians who are part of the orchestra does not depend on the success of this project to make a living. & David Grimal admits that Les Dissonaces represents "a luxury", stating that none of the musicians who are part of the orchestra does not depend on the success of this project for living. & David Grimal admits Les Dissonaces represent "a luxury," saying none of the musicians who make up the orchestra depinde of the success of this project to live. \\\hline
\end{tabular}
%}
\caption{Top 10 segments by $|\delta_{\bleu}(i, h_S, h_U)|$ on RO-EN.}
\label{tab:bleu-top-10-roen}
\end{table*}

\end{document}